\documentclass{valence} 
\usepackage{soul} 
\usepackage{float}
\usepackage{amsmath}
\usepackage{amsfonts}
\usepackage{subcaption}
\usepackage{longtable}

\usepackage[export]{adjustbox}
\usepackage{comment}
\usepackage{mathtools}
\usepackage{authblk}
\usepackage{fancyhdr}
\usepackage{array}
\usepackage{hyperref}
\usepackage{xspace}
\usepackage{microtype}
\usepackage{graphicx}
\usepackage{makecell}  
\usepackage{booktabs, threeparttable, multirow}

\usepackage{amssymb}
\usepackage{amsthm}
\usepackage{enumitem}
\usepackage{subcaption}
\usepackage{graphicx}
\usepackage[capitalize,noabbrev]{cleveref}
\usepackage[inkscapelatex=false]{svg}
\usepackage{ulem}

\usepackage{amsmath,amsfonts,bm}

\def\eqref#1{equation~\ref{#1}}

\def\1{\bm{1}}

\DeclareMathAlphabet{\mathsfit}{\encodingdefault}{\sfdefault}{m}{sl}
\SetMathAlphabet{\mathsfit}{bold}{\encodingdefault}{\sfdefault}{bx}{n}

\DeclareMathOperator*{\argmax}{arg\,max}

\newcommand{\modelname}{SAFE-T}

\title{Conditional Chemical Language Models are Versatile Tools in Drug Discovery}

\setrunningtitle{Conditional CLMs are Versatile Tools in Drug Discovery}

\author[1]{Lu Zhu}
\author[1,\star]{Emmanuel Noutahi}

\affiliation[1]{Valence Labs, Recursion}
\contribution[\star]{\small Correspondence: emmanuel@valencelabs.com}

\abstract{
Generative chemical language models (CLMs) have demonstrated strong capabilities in molecular design, yet their impact in drug discovery remains limited by the absence of reliable reward signals and the lack of interpretability in their outputs. We present \modelname, a generalist chemical modeling framework that conditions on biological context---such as protein targets or mechanisms of action---to prioritize and design molecules without relying on structural information or engineered scoring functions. 
\modelname~models the conditional likelihood of fragment-based molecular sequences given a biological prompt, enabling principled scoring of molecules across tasks such as virtual screening, drug--target interaction prediction, and activity cliff detection. Moreover, it supports goal-directed generation by sampling from this learned distribution, aligning molecular design with biological objectives. In comprehensive zero-shot evaluations across predictive (LIT-PCBA, DAVIS, KIBA, ACNet) and generative (DRUG, PMO) benchmarks, \modelname~consistently achieves performance comparable to or better than existing approaches while being significantly faster. Fragment-level attribution further reveals that \modelname~captures known structure--activity relationships, supporting interpretable and biologically grounded design. Together with its computational efficiency, these results demonstrate that conditional generative CLMs can unify scoring and generation to accelerate early-stage drug discovery.}

\begin{document}
\maketitle

\section{Introduction}
\label{sec:intro}

Generative chemical language models (CLMs) have emerged as powerful tools for molecular design, enabling automated exploration of chemical space through deep learning architectures such as RNNs, VAEs, and Transformers~\citep{grisoni2023chemical, gomez2018automatic, moret2023leveraging, loeffler2024reinvent, segler2018generating, irwin2022chemformer}. By learning the implicit rules of chemical structure and reactivity from large molecular datasets, CLMs can generate drug-like compounds that satisfy multiple design constraints simultaneously. This capability is particularly valuable in early-stage drug discovery, where rapid and efficient navigation of vast chemical spaces is critical for identifying novel candidates. 

Despite this promise, CLMs face two fundamental limitations in real-world applications. First, the lack of reliable reward functions hinders optimization toward biological objectives~\citep{wognum2024call, tossou2024real}, especially in hit-to-lead and target discovery phases where biological complexity and limited feedback dominate. Even with the known limitations of traditional virtual screening approaches---such as chemical promiscuity and low novelty~\citep{lyu2019ultra, paggi2024art}---CLMs often struggle to outperform them due to this lack of biological grounding. Second, most CLMs operate as black boxes, offering little interpretability to guide structure--activity relationship (SAR) analysis~\citep{wellawatte2022model, jimenez2020drug}, limiting their adoption in medicinal chemistry workflows.

Recent advances in high-throughput assays have produced rich biological annotations describing how compounds modulate protein targets and mechanisms of action~\citep{davis2020mechanism, subramanian2017next}. However, molecular design tools rarely take full advantage of this biological context. While many ML methods focus on predicting ligand--target affinity~\citep{shi2024review}, they often overlook broader questions of downstream function, engagement, and pathway-level impact~\citep{copeland2006drug, kenakin2019biased, swinney2011were, wacker2017crystal, lin2019off, shen2023allosteric}. This disconnect presents an opportunity for more biologically grounded approaches to molecule design.

\begin{figure*}[tb]
    \begin{center}
\centerline{\includegraphics[width=1\textwidth]{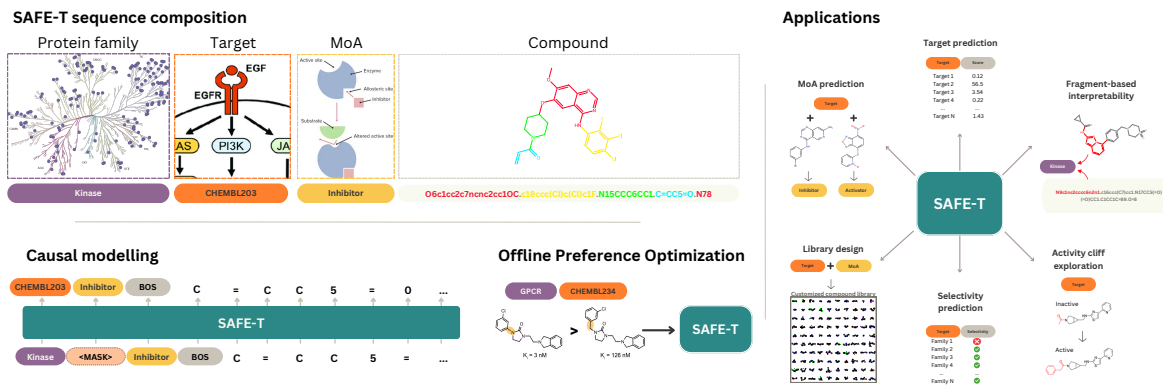}}   
    \caption{Overview of the \modelname~framework for biologically informed molecular design. \modelname~integrates SAFE strings with biological context tokens that encode target proteins and mechanisms of action. Through a three-stage training process (pretraining, biological context fine-tuning, and preference optimization), it learns associations between molecular fragments and biological activities. This enables \modelname~to generalize across multiple drug discovery tasks in zero-shot settings.} \label{fig:illustration}
\end{center}
\vskip -0.2in
\end{figure*}

We introduce \textbf{\modelname} (\textbf{S}equential \textbf{A}ttachment-based \textbf{F}ragment \textbf{E}mbedding with \textbf{T}arget-conditioning), a conditional generative framework trained to model the likelihood of molecules given biological prompts, such as protein targets, mechanisms of action (MoA), or cellular annotations (see \autoref{fig:illustration}). 

\modelname~unifies biologically conditioned scoring and generation in a single model. It achieves strong zero-shot performance across diverse drug discovery tasks, including  drug-target interaction prediction, molecular prioritization, and activity cliff detection.  It also supports interpretable design, with fragment-level attributions that align with established structure--activity relationships. Together, these capabilities make \modelname~a practical and scalable framework for early-stage drug discovery, especially in the abscence of reliable reward functions.

\section{Related Work} 
\label{sec:relatedwork}

We position \modelname~at the intersection of three key areas of prior work: fragment-based representations, conditional generation, and interpretability.

\subsection{Fragment-Constrained Design and SAFE Representations}

Fragment-based molecular design plays a central role in medicinal chemistry, particularly in hit-to-lead optimization, where systematic modification of privileged scaffolds informs structure--activity relationships~\citep{Jhoti2013, Woodhead2024}. However, most standard line notations, such as SMILES, lack explicit representations of fragment structure and connectivity, limiting their interpretability and controllability for modular design.

To address these limitations, recent work has proposed a range of fragment-level representations for substructure-aware design~\citep{jin2020hierarchical, wu2024t, mastrolorito2024fragsmiles, noutahi2024gotta}. Among them, Sequential Attachment-Based Fragment Embedding (SAFE)~\citep{noutahi2024gotta,mesbahi2024safe} strikes a practical balance between chemical fidelity and usability: it represents molecules as sequences of chemically connected fragments in a textual format compatible with SMILES-based infrastructure. SAFE preserves the generation reliability of SMILES and SELFIES while offering improved alignment with synthetic feasibility and enabling structure-constrained generation. Compared to other fragment-based approaches, it maintains a simple and interpretable format that supports both substructure- and pattern-based design, and flexibly handles single- or multi-fragment constraints. These properties make it particularly well suited for conditional generative modeling, where precision and modularity are essential. \modelname~leverages SAFE to enable fragment-level generation and attribution grounded in biological context.

\subsection{Conditional Generation and Generalist Architectures}

Conditional generation in molecular design has evolved from property-driven control to more pharmacologically relevant conditioning. Early work demonstrated property-guided generation by conditioning on numerical endpoints such as logP, QED, or molecular weight~\citep{kang2018conditional, li2018multi, kotsias2020direct}. More recently, transformer-based models such as cMolGPT~\citep{wang2023cmolgpt} and Llamol~\citep{dobberstein2024llamol} have extended this paradigm through token-level conditioning strategies that support generation from multi-property or multi-condition prompts.

To go beyond physicochemical control, several studies have proposed conditioning molecule generation on biological data. This includes target-specific design guided by protein structures~\citep{skalic2019shape, imrie2021deep, xu2021novo, ragoza2022generating, wu2024tamgen, lin2024functional, qian2024kgdiff, liu2022generating, feng2024generation, jiang2024pocketflow}, gene expression profiles~\citep{mendez2020novo}, and phenotypic data~\citep{pham2022fame}. While promising, these approaches tend to be narrowly scoped: they focus on single data modalities, require structural input, or produce compounds with limited synthetic feasibility~\citep{imrie2021deep, liu2022generating}. Their designs are also typically tailored to specific tasks, limiting generalization across biological contexts.

In parallel, large-scale generalist models have begun to unify molecular and biological representations for therapeutic tasks. For example, TxGemma~\citep{wang2025txgemma}, a family of open models derived from Gemma-2, uses large language models trained on therapeutic datasets to predict molecular properties and facilitate conversational interaction with biological data. Although efficient for molecular property prediction and data analysis, TxGemma and related LLM-derived models are not designed for fragment-constrained or bioactivity-driven molecular generation.

\modelname~offers an alternative and integrative approach: it unifies biologically conditioned scoring and generation in a single framework, trained end-to-end on fragment-level sequences and biological prompts. Unlike most conditional generators, \modelname~typically does not require target-specific fine-tuning. Unlike predictive-only models, it enables the design of new molecules aligned with biological objectives. As interest in generalist therapeutic models continues to grow, biologically grounded generative frameworks like \modelname~may serve as foundations for the next generation of therapeutic models that integrate biological reasoning with controllable molecular design.

\subsection{Explainability and Interpretability in Molecular Modeling}

The ability to interpret how molecular structures influence biological activity is fundamental to rational drug design. Attribution techniques such as SHAP~\citep{lundberg2017unified} and LIME~\citep{ribeiro2016should} have been used to map predictions to specific structural features, while counterfactual approaches have explored local structure--activity landscapes through minimal molecular modifications~\citep{wellawatte2022model}. Recent work has also demonstrated the utility of attention mechanisms, originally developed for natural language processing~\citep{vig2019analyzing}, in analyzing chemical transformer models~\citep{hodl2023explainability}. However, while tools exist for interpreting generative language models~\citep{sarti2023inseq}, their adaptation to molecular generation tasks remains limited—particularly for conditional generation, where tracing the relationship between conditioning context and molecular output is critical.

\section{\modelname~Framework}
\label{sec:method}

We introduce \textbf{\modelname} (\textbf{S}equential \textbf{A}ttachment-based \textbf{F}ragment \textbf{E}mbedding with \textbf{T}arget-conditioning), a unified framework for biologically conditioned molecular design and scoring (see \autoref{fig:illustration}). \modelname~is designed to address two central challenges in drug discovery: the absence of robust reward functions for guiding generative models, and the disconnect between data-driven design and biological relevance. Instead of optimizing proxy scores, \modelname~learns to condition molecular generation directly on high-level biological prompts, enabling unified molecule generation, prioritization, and scoring within a single probabilistic model.

\subsection{Conditional Generative Modeling}

Each training sample consists of a molecule \(x = (x_1, \dots, x_T)\), represented in SAFE format using $T$ tokens, and a biological condition \(c = (c_{\text{fam}}, c_{\text{tgt}}, c_{\text{moa}})\). These components specify the target protein family (e.g., ``Kinase''), the specific protein target (e.g., ``EGFR''), and the compound's mechanism of action (e.g., ``inhibitor''), respectively. Together, they encode pharmacological intent at multiple levels of abstraction.

To supervise training, we compile MoAT-DB (see Appendix~\ref{appendix:dataset}), a dataset of molecule–context pairs aggregated from DrugBank \citep{Wishart2018}, Connectivity Map \citep{subramanian2017next}, the Therapeutic Target Database \citep{Zhou2024}, and ChEMBL34 \citep{Zdrazil2024}. Each triplet \(c\) is prepended to the molecular sequence, and the model learns the conditional likelihood:
\[
p_\theta(x \mid c) = \prod_{t=1}^T p_\theta(x_t \mid x_{<t}, c),
\]
where \(p_\theta\) is a transformer-based autoregressive language model. The dataset is of the form:
\[
\mathcal{D} = \{(c^i, x^i) \mid c^i = (c^i_{\text{fam}}, c^i_{\text{tgt}}, c^i_{\text{moa}}),\; x^i = (x^i_1, \ldots, x^i_T) \}.
\]
\modelname~supports several core tasks in early-stage drug discovery: (i) conditional molecule generation aligned with therapeutic goals, (ii) prioritization of candidate compounds based on biological prompts, and (iii) fragment-level interpretability for understanding structure–activity relationships. These capabilities all emerge from the same learned conditional distribution, trained in three stages outlined below and detailed in Appendix~\ref{appendix:training_processes}.

\paragraph{Stage 1: Pretraining on Chemical Structure.}
We first pretrain the model on a large corpus of unlabeled lead-like molecular sequences, using a null biological context in which all condition tokens are masked:
\[
c^\text{null} = (\text{\textless mask\textgreater}, \text{\textless mask\textgreater}, \text{\textless mask\textgreater})
\]
During this stage, the model learns to generate molecules by modeling $p_\theta(x \mid c^\text{null})$ capturing chemical syntax and the principles of fragment assembly in the SAFE representation. This chemically grounded initialization ensures that the model produces valid and syntactically coherent molecules, which is especially important when structural constraints are imposed or when generalizing to novel biological prompts in downstream tasks.

\paragraph{Stage 2: Fine-tuning with Biological Context.}
We then fine-tune the model on paired molecule--context data \((x, c)\), drawn from MoAT-DB.  This teaches the model to associate structural patterns with therapeutic goals. To improve generalization and robustness to partial knowledge, random subsets of \(c\) are masked during training. At inference time, the model can flexibly condition on full or partial prompts (e.g., target family only), enabling context-aware generation even in settings where only limited biological information is available.

\paragraph{Stage 3: Preference Calibration.}
Finally, we calibrate the model's likelihoods for prioritization tasks using paired comparisons \((c, x^+, x^-)\), where \(x^+\) is preferred over \(x^-\) under condition \(c\). The model is trained to increase \(p_\theta(x^+ \mid c)\) relative to \(p_\theta(x^- \mid c)\), improving its ability to reflect compound preference in virtual screening or activity cliff detection. Details of this training stage are provided in Appendix~\ref{appendix:training_processes}.

\subsection{Predictive and Interpretability Tasks}
\label{sec:predictive_tasks}

Although \modelname~is trained as a conditional generative model, its learned distribution \(p_\theta(x \mid c)\) can be repurposed for several key predictive and interpretability tasks central to early-stage drug discovery. These capabilities emerge naturally from the structure of the model.

\paragraph{Predictive Modeling.}
We formulate prediction as inference over the biological condition \(c\) given a molecule \(x\). Specifically, for any unknown subset of \(c\), we score candidate completions using the joint posterior:
\[
p(c \mid x) \propto p_\theta(x \mid c) \cdot p(c)
\]
This formulation unifies tasks such as drug-target interaction (DTI) prediction, MoA classification, and target family assignment under a single probabilistic interface. It also supports ranking compounds within a given context, enabling zero-shot molecular prioritization and selectivity profiling. Detailed mathematical and implementation information are provided in Appendix~\ref{appendix:predictive_task}.

\paragraph{Fragment-level Attribution.}
SAFE's fragment-based structure allows us to approximate the influence of individual substructures on model behavior. For a given fragment \(f_i \in x\), we estimate its contribution by comparing the log-likelihood of the original molecule to that of a counterfactual variant \(x_{f_i'}\) in which \(f_i\) is replaced:
\[
\psi_{f_i} = \log p_\theta(x \mid c) - \mathbb{E}_{f_i'} \left[\log p_\theta(x_{f_i'} \mid c)\right]
\]
This contrastive attribution provides insights into how structural components modulate predicted fitness under a biological prompt. While not a mechanistic explanation, this offers actionable signal for SAR-driven design. Higher-order interactions and counterfactual modifications are explored in Appendix~\ref{appendix:sar-analysis}.

\section{Results}
\label{sec:results}
We evaluate \modelname~across a broad range of generative, predictive, and interpretability tasks. Experiments assess its ability to generate valid and structurally diverse molecules, prioritize compounds in zero-shot settings, generalize to unseen targets, and uncover fragment-level structure–activity relationships. Full implementation details, training settings, and additional analyses are provided in Appendix~\ref{sec:appendix}.

\subsection{Experimental setup}

\modelname~is trained on MoAT-DB, our curated dataset of compound--target--MoA triplets constructed.  To assess generalization, we designed three held-out test sets: (i) $\mathcal{D}_\text{val}$, an IID validation set; (ii) $\mathcal{D}_\text{Mol}$, containing novel molecules unseen during training; and (iii) $\mathcal{D}_\text{MoA}$, featuring unseen Target-MoA combinations. Details on dataset construction are provided in Appendix~\ref{appendix:dataset}. We train multiple \modelname~variants at different model sizes (Appendix~\ref{appendix:training_processes}) and evaluate them across key tasks. Generative modeling performance was evaluated using standard metrics, including Validity, internal diversity (Int.Div), Synthetic Accessibility Score (SAS), Drug-Likeness (QED), and Constraint Preservation (Cons.Pres), while predictive tasks were assessed using ROC-AUC, Top-1\% Accuracy (ACC@1\%), and Enrichment Factor (EF@1\%). The rest of this section presents our main findings across generation, prediction, prioritization, and interpretability. All results are based on the largest \modelname~variant unless otherwise noted.

\subsection{Molecular Generation Performance}
\begin{table*}[bt]
\centering
\caption{Performance of large \modelname~on six molecule generation tasks against a non-conditional baseline model. Metrics include Synthetic Accessibility Score (SAS), Quantitative Estimate of Drug-Likeness (QED), Internal Diversity (Int.Div), Validity, and Constraint Preservation (Cons.Pres). The best performance for each metric is highlighted in bold.}
\label{tab:\modelname-mol-gen}
\resizebox{\linewidth}{!}{
\begin{tabular}{llccccc}
\toprule
$\mathbf{Task}$ & $\mathbf{Setup}$ & $\mathbf{\downarrow SAS}$ & $\mathbf{\uparrow QED}$ & $\mathbf{\uparrow Int.Div (\%)}$ & $\mathbf{\uparrow Validity (\%)}$ & $\mathbf{\uparrow Cons.Pres (\%)}$ \\
\midrule
\multirow{3}{*}{De novo}   
& SAFE (baseline)  & $\mathbf{2.66 \pm  0.71}$      & $\mathbf{0.60 \pm 0.19}$       & $87.12$                & $99.16$                & -                         \\
& \modelname~(masked)  & $2.72 \pm 0.72$  & $0.59 \pm 0.19$  & $\mathbf{87.42}$  & $99.13$  & -- \\
& \modelname~(context) & $2.77 \pm 0.37$  & $0.52 \pm 0.11$  & $82.42 \pm 2.20$  & $\mathbf{99.38 \pm 0.67}$  & -- \\
\midrule
\multirow{3}{*}{\makecell[l]{Linker\\Prediction}}   
& SAFE (baseline)             & $3.24 \pm 0.90$      & $0.58 \pm 0.15$        & $\mathbf{60.56 \pm 9.95}$       & $\mathbf{100.00 \pm 0.00}$      & $\mathbf{100.00 \pm 0.00}$               \\
& \modelname~(masked) & $3.22 \pm 0.94$  & $\mathbf{0.60 \pm 0.18}$  & $57.09 \pm 13.90$  & $\mathbf{100.00 \pm 0.00}$  & $\mathbf{100.00 \pm 0.00}$ \\
& \modelname~(context) & $\mathbf{3.14 \pm 0.98}$  & $0.58 \pm 0.17$  & $59.11 \pm 10.61$  & $\mathbf{100.00 \pm 0.00}$  & $\mathbf{100.00 \pm 0.00}$ \\
\midrule
\multirow{3}{*}{\makecell[l]{Scaffold\\Decoration}} 
& SAFE (baseline)            & $\mathbf{3.21 \pm 0.93}$      & $\mathbf{0.50 \pm 0.16}$         & $44.98 \pm 16.89$     & $98.92 \pm 3.83$       & $99.78 \pm 1.00$                \\
& \modelname~(masked)  & $3.24 \pm 0.93$  & $0.45 \pm 0.11$  & $43.95 \pm 11.95$  & $98.63 \pm 5.03$  & $\mathbf{100.00 \pm 0.00}$ \\
& \modelname~(context) & $3.33 \pm 0.84$  & $0.46 \pm 0.14$  & $\mathbf{45.86 \pm 12.79}$  & $\mathbf{99.60 \pm 0.81}$  & $99.95 \pm 0.22$ \\
\midrule
\multirow{3}{*}{\makecell[l]{Scaffold\\morphing}}
& SAFE (baseline)  & $3.60 \pm 0.91$    & $\mathbf{0.61 \pm 0.02}$ & $60.80 \pm 9.60$          & $\mathbf{100.00 \pm 0.00}$ & $\mathbf{100.00 \pm 0.00}$ \\
& \modelname~(masked)  & $3.47 \pm 1.17$          & $0.54 \pm 0.17$          & $\mathbf{61.67 \pm 11.14}$ & $\mathbf{100.00 \pm 0.00}$ & $\mathbf{100.00 \pm 0.00}$ \\
& \modelname~(context) & $\mathbf{3.21 \pm 0.94}$ & $0.59 \pm 0.16$          & $59.88 \pm 10.47$          & $\mathbf{100.00 \pm 0.00}$ & $\mathbf{100.00 \pm 0.00}$ \\
\midrule
\multirow{3}{*}{\makecell[l]{Motif\\extension}}
& SAFE (baseline)  & $3.75 \pm 0.65$            & $0.57 \pm  0.08$         & $\mathbf{68.10 \pm 8.90}$  & $\mathbf{100.00 \pm 0.00}$ & $\mathbf{100.00 \pm 0.00}$ \\
&\modelname~(masked) & $3.31 \pm 0.69$ & $\mathbf{0.60 \pm 0.17}$ & $58.86 \pm 11.16$ & $99.11 \pm 1.18$ & $100.00 \pm 0.00$ \\
&\modelname~(context) & $\mathbf{3.18 \pm 0.75}$ & $0.59 \pm 0.18$& $61.64 \pm 9.80$ & $98.86 \pm 2.40 $& $100.00 \pm 0.00$ \\
\midrule
\multirow{3}{*}{\makecell[l]{Superstructure}}   
& SAFE (baseline)  & $3.87 \pm 0.09$           & $0.43 \pm 0.02$          & $\mathbf{71.50 \pm 5.90}$  & $\mathbf{100.00 \pm 0.00}$ & $\mathbf{100.00 \pm 0.00}$ \\
& \modelname~(masked) &	$\mathbf{3.48 \pm 0.94}$ &	$\mathbf{0.50 \pm 0.15}$ &	$65.51 \pm 7.38$ &	$98.77 \pm 4.00$ &	$99.81 \pm 0.00$ \\
& \modelname~(context) &	$3.58 \pm 0.92$ &	$\mathbf{0.50 \pm 0.15}$ &	$67.26 \pm 6.08$ &	$97.29 \pm 5.23$ &	$99.42 \pm 0.00$ \\
\bottomrule
\end{tabular}
}
\end{table*}
While SAFE-based models have demonstrated strong generative performance in previous work \citep{lee2025genmol,noutahi2024gotta}, our primary focus here is to evaluate whether the integration of biological context conditioning maintains these capabilities. Therefore, we assess performance on both \textit{de novo} generation and structure-constrained tasks (linker design, scaffold decoration) using the DRUG benchmark \citep{noutahi2024gotta}. For each experiment, we generated 10,000 molecules under different conditions: masked biological context (equivalent to unconditional generation) and true biological contexts for structure-constrained tasks (or random contexts for \textit{de novo} generation). Examples of generated molecules are presented in \autoref{appendix:fig:sample-molecules}.

\autoref{tab:\modelname-mol-gen} shows that \modelname~maintains performance comparable to a non-conditional baseline even after biological context fine-tuning, with high validity ($>95\%$) across all generation modes. In structure-constrained tasks, it achieves a near-perfect structural constraint satisfaction ($>99\%$) while maintaining high synthetic accessibility, regardless of whether biological contexts are masked or specified. However, we note a decrease in diversity that is typical of structure-conditioned generation. These results demonstrate that a single \modelname~model can seamlessly handle biological context conditioning, structural constraints, or both simultaneously without task-specific adjustments. We provide detailed performance comparisons across training stages and model sizes in Appendix~\ref{appendix:ablation}.

\subsection{Zero-shot Molecular Optimization}
\label{sec:goal-directed-opt}

\begin{table*}[tb]
    \centering
\caption{Performance comparison of \modelname~against state-of-the-art methods in goal-directed molecular optimization benchmarks (DRD2, JNK3, GSK-3$\beta$). Metrics shown are the mean and standard deviation of PMO AUC (top-10). Results for baseline methods were obtained from \citep{lee2025genmol} based on benchmarks from \citep{Gao2022}. \modelname~demonstrates consistently improved performance across all evaluated targets.}
    \label{tab:oracle_performance}
    \begin{tabular}{lccc}
        \toprule
        \textbf{Method} & \textbf{DRD2} & \textbf{JNK3} & \textbf{GSK-3$\beta$} \\
        \midrule
        \textbf{\modelname~(Ours)} & $\mathbf{0.9950 \pm 0.0000}$ & $\mathbf{0.9927 \pm 0.0003}$ & $\mathbf{0.9948 \pm 0.0001}$ \\
        GenMol & $\mathbf{0.9950 \pm 0.0000}$ & $0.8560 \pm 0.0160$ & $0.9860 \pm 0.0030$ \\
        f-RAG & $0.9920 \pm 0.0000$ & $0.9040 \pm 0.0040$ & $0.9690 \pm 0.0030$ \\
        GeneticGFN & $0.9740 \pm 0.0060$ & $0.7640 \pm 0.0690$ & $0.8810 \pm 0.0420$ \\
        MolGA & $0.9360 \pm 0.0160$ & $0.7020 \pm 0.1230$ & $0.8430 \pm 0.0390$ \\
        REINVENT & $0.9450 \pm 0.0070$ & $0.7830 \pm 0.0230$ & $0.8650 \pm 0.0430$ \\
        GraphGA & $0.9640 \pm 0.0120$ & $0.5530 \pm 0.1360$ & $0.7880 \pm 0.0700$ \\
        SELFIES-REINVENT & $0.9430 \pm 0.0050$ & $0.6310 \pm 0.0640$ & $0.7800 \pm 0.0370$ \\
        GPBO & $0.9230 \pm 0.0170$ & $0.5640 \pm 0.1550$ & $0.8510 \pm 0.0410$ \\
        STONED & $0.9130 \pm 0.0200$ & $0.5230 \pm 0.0920$ & $0.6680 \pm 0.0490$ \\
        LSTM HC & $0.9190 \pm 0.0150$ & $0.6610 \pm 0.0390$ & $0.8390 \pm 0.0150$ \\
        SMILES-GA & $0.9080 \pm 0.0190$ & $0.3160 \pm 0.0220$ & $0.6290 \pm 0.0440$ \\
        SynNet & $0.9690 \pm 0.0040$ & $0.6300 \pm 0.0340$ & $0.7890 \pm 0.0320$ \\
        DoG-Gen & $0.9480 \pm 0.0010$ & $0.5950 \pm 0.0230$ & $0.8310 \pm 0.0210$ \\
        MARS & $0.8910 \pm 0.0200$ & $0.4890 \pm 0.0950$ & $0.5520 \pm 0.0370$ \\
        MIMOSA & $0.7990 \pm 0.0170$ & $0.3600 \pm 0.0630$ & $0.5540 \pm 0.0420$ \\
        MolPal & $0.7830 \pm 0.0090$ & $0.3390 \pm 0.0090$ & $0.5550 \pm 0.0110$ \\
        SELFIES-LSTM HC & $0.7290 \pm 0.0340$ & $0.2070 \pm 0.0130$ & $0.4230 \pm 0.0180$ \\
        DoG-AE & $0.9430 \pm 0.0090$ & $0.4690 \pm 0.1380$ & $0.6010 \pm 0.0910$ \\
        GFlowNet & $0.5900 \pm 0.0700$ & $0.4400 \pm 0.0220$ & $0.6510 \pm 0.0260$ \\
        SELFIES-VAE BO & $0.5690 \pm 0.0390$ & $0.2080 \pm 0.0220$ & $0.3500 \pm 0.0340$ \\
        Screening & $0.5450 \pm 0.0150$ & $0.2380 \pm 0.0240$ & $0.4380 \pm 0.0340$ \\
        SMILES-VAE BO & $0.5550 \pm 0.0430$ & $0.2410 \pm 0.0260$ & $0.3860 \pm 0.0060$ \\
        GFlowNet-AL & $0.4680 \pm 0.0460$ & $0.3620 \pm 0.0210$ & $0.5880 \pm 0.0150$ \\
        JT-VAE BO & $0.5060 \pm 0.1360$ & $0.2220 \pm 0.0090$ & $0.3500 \pm 0.0510$ \\
        Graph MCTS & $0.3000 \pm 0.0500$ & $0.1100 \pm 0.0190$ & $0.2810 \pm 0.0220$ \\
        MolDQN & $0.0250 \pm 0.0010$ & $0.1110 \pm 0.0080$ & $0.2410 \pm 0.0080$ \\
        \bottomrule
    \end{tabular}
\end{table*}

To assess its capacity for goal-directed molecular design without task-specific training, we tested \modelname~on three standard optimization tasks targeting DRD2 \citep{olivecrona2017molecular}, JNK3 \citep{li2018multi}, and GSK-3$\beta$, using the Practical Molecular Optimization (PMO) benchmark \citep{Gao2022}. This benchmark quantifies how efficiently a model finds top compounds using the area under the curve (AUC) of the best 10 oracle scores from 10,000 generated molecules.

Unlike most approaches that rely on fine-tuning or reinforcement learning, \modelname~employs a purely generative strategy driven by conditional prompts, requiring no optimization for specific tasks. Remarkably, as shown in \autoref{tab:oracle_performance}, this zero-shot methodology achieves state-of-the-art performance on all three benchmark tasks outperforming specialized methods (e.g., GenMol, REINVENT) that uses explicit optimization strategies.

Furthermore, \modelname~generates chemically diverse and synthetically accessible molecules (see \autoref{tab:pmo_properties} in Appendix~\ref{appendix:ablation}), indicating it produces not just high-scoring, but lead-like molecules.

This demonstrates \modelname's ability to infer structure–activity relationships from biological context alone, enabling effective zero-shot design without optimization. Importantly, our results suggest that conditional generation with generalist models offers a powerful alternative to established reward-based or search-centric drug design paradigms.

\subsection{Zero-shot Prediction of Target Interaction and Target Family Selectivity}
\label{subsec:target_family_selectivity}

Predicting drug selectivity and polypharmacology is crucial for designing effective therapeutics and minimizing off-target effects. We evaluate \modelname's ability to predict both overall target family affinities for broader interaction trend and specific drug-target interactions (DTI). While detailed DTI results are provided in Appendix~\ref{appendix:dti-prediction}, this section focuses on target family-level predictions.

We evaluate \modelname~on the MoAT-DB test sets ($\mathcal{D}_\text{val}$, $\mathcal{D}_\text{Mol}$, $\mathcal{D}_\text{MoA}$) and the kinase-inhibitor-specific DAVIS \citep{davis2011comprehensive} and KIBA \citep{tang2014making} datasets. For target family prediction, individual targets are grouped into their respective families. This reformulation aligns with real-world drug discovery, where kinase inhibitors often exhibit family-wide activity rather than strict target selectivity.  To ensure a fair evaluation, all Compound--Target pairs present in the training data were removed. \modelname's performance is reported using the $l_\text{null}$ normalization technique (Appendix~\ref{appendix:lkl_normalization}).

As shown in \autoref{tab:tgt_fam_moa_pred}, \modelname~generalizes well across target families, outperforming a supervised baseline (linear probing of MolGPS, a molecular graph foundation model \citep{Sypetkowski2024}) on most out-of-distribution test sets. It achieves strong ROC-AUC scores on the kinase datasets (DAVIS: 0.706, KIBA: 0.759) and unseen Target--MoA combinations ($\mathcal{D}_\text{MoA}$: 0.892), suggesting that it captures meaningful family-level interaction patterns. However, it underperforms on the IID evaluation set ($\mathcal{D}_\text{val}$: 0.777 vs. 0.942), likely due to a higher proportion of promiscuous compounds with broad target profiles, which may be better captured by representation-focused models like MolGPS.  

Similarly, \autoref{appendix:tab:target_prediction} demonstrates \modelname’s competitive performance on DTI prediction. This consistent performance across both family-level and individual target predictions suggests that \modelname~effectively learns a hierarchical representation of target-ligand interactions, allowing it to generalize across different levels of target specificity. Interestingly, the preference-tuned \modelname$_\text{post}$ maintains similar performance in target family prediction, indicating that preference optimization does not compromise its ability to model biological selectivity.

\begin{table*}[tb]
\centering
\caption{Target family - MoA prediction on MoAT-DB testset, DAVIS and KIBA datasets}
\label{tab:tgt_fam_moa_pred}
\begin{tabular}{llcc}
\toprule
\textbf{Dataset} & \textbf{Method} & \textbf{ROC-AUC} & \textbf{PR-AUC} \\
\midrule
\multirow{3}{*}{$\mathcal{D}_\text{val}$} 
    & MolGPS+MLP               & \textbf{0.942} & \textbf{0.524} \\
    & \modelname~              & 0.777          & 0.286 \\
    & \modelname\textsubscript{post} & 0.749          & 0.243 \\
\midrule
\multirow{3}{*}{$\mathcal{D}_\text{Mol}$} 
    & MolGPS+MLP               & 0.682          & 0.224 \\
    & \modelname~              & \textbf{0.786} & \textbf{0.283} \\
    & \modelname\textsubscript{post} & 0.745          & 0.233 \\
\midrule
\multirow{3}{*}{$\mathcal{D}_\text{MoA}$} 
    & MolGPS+MLP               & 0.684          & 0.231 \\
    & \modelname~              & \textbf{0.892} & \textbf{0.339} \\
    & \modelname\textsubscript{post} & 0.839          & 0.242 \\
\midrule
\multirow{3}{*}{DAVIS}     
    & MolGPS+MLP               & 0.686          & \textbf{0.801} \\
    & \modelname~              & \textbf{0.706} & 0.778 \\
    & \modelname\textsubscript{post} & \textbf{0.705} & 0.756 \\
\midrule
\multirow{3}{*}{KIBA}      
    & MolGPS+MLP               & 0.706          & 0.633 \\
    & \modelname~              & \textbf{0.759} & \textbf{0.687} \\
    & \modelname\textsubscript{post} & 0.738          & 0.653 \\
\bottomrule
\end{tabular}
\end{table*}

\subsection{Zero-shot Performance on Virtual Screening Benchmarks}
\label{section:mol_prioritization}

We evaluated \modelname~on molecule prioritization using several standard benchmarks, focusing on the LIT-PCBA dataset \citep{tran2020lit}, which presents an exceptionally challenging benchmark of 15 targets with stringently curated actives and inactives from dose-response assays. Results on LIT-PCBA are reported here due to its minimal overlap with MoAT-DB. Detailed dataset descriptions, evaluation metrics, and additional results are provided in Appendix~\ref{appendix:mol_prioritization}. \modelname’s performance is reported using the $l_\text{pop}$ normalization technique (Appendix~\ref{appendix:lkl_normalization}).

\begin{figure}[bt]
    \begin{center}
\centerline{\includegraphics[width=0.7\linewidth]{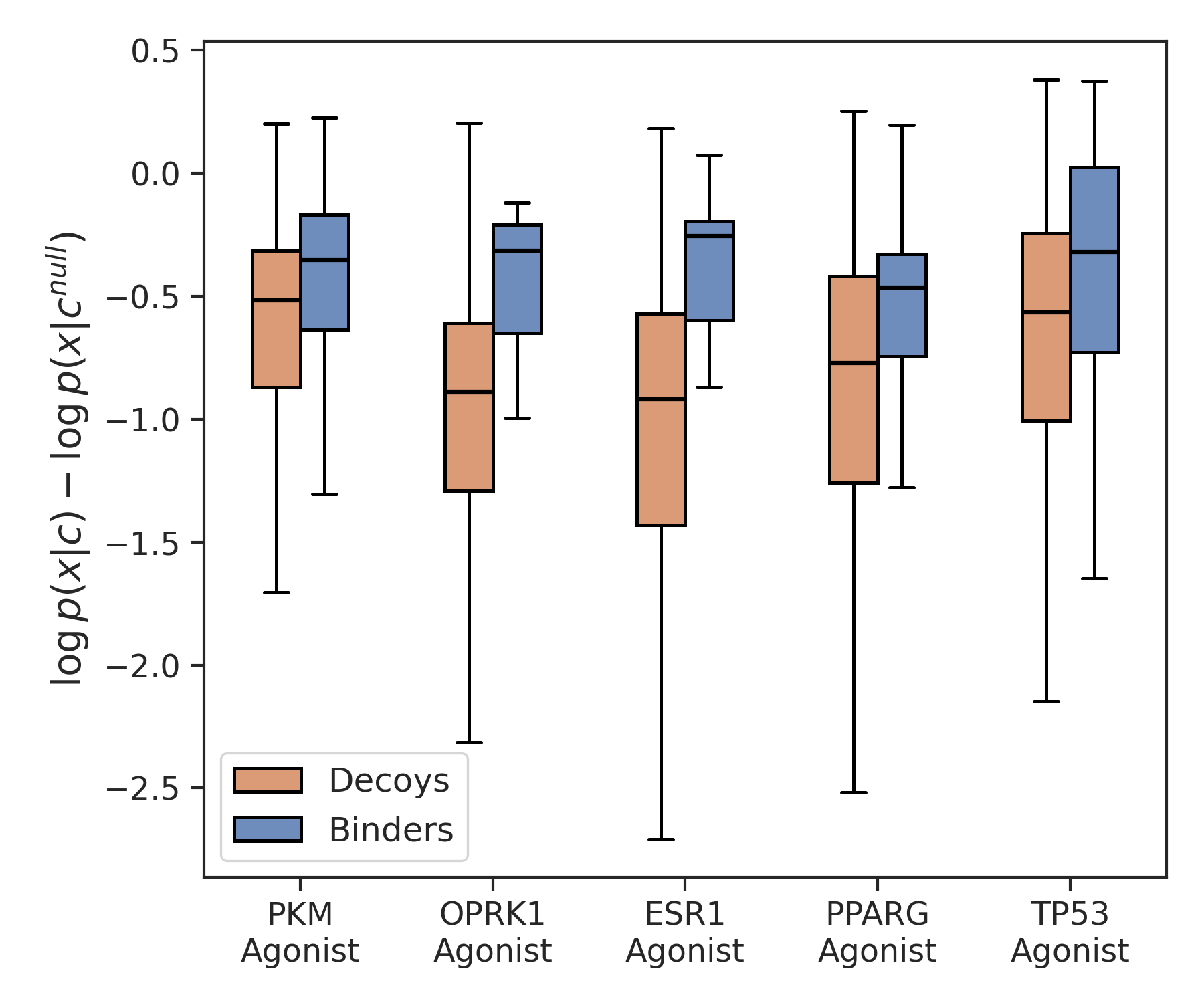}}
\caption{Normalized likelihood distributions for decoys and binders on selected target from the LIT-PCBA benchmark, showing \modelname's enrichment of active compounds.}
\label{fig:pcba_distr}
\end{center}
\vskip -0.2in
\end{figure}

As shown in \autoref{tab:pcba-gpcr}, \modelname~achieves strong zero-shot performance on LIT-PCBA (EF@1\%: 6.63), outperforming both traditional structure-based methods like Glide-SP~\citep{halgren2004glide} (EF@1\%: 3.41) and recent deep learning approaches such as DrugCLIP~\citep{gao2024drugclip} (EF@1\%: 5.51). While some overlap exists between active drug-target pairs and our training data (common in virtual screening evaluations where scoring functions are trained on large datasets like PDBBind and BindingDB), \modelname’s ability to enrich active compounds without structural information or task-specific fine-tuning is particularly noteworthy (Fig.~\ref{fig:pcba_distr}). This advantage likely stems from its ability to leverage MoA information, which the baseline methods typically ignore.

\paragraph{Computational Cost.} 

\modelname~enables high-throughput virtual screening, with GPU inference speeds reaching nearly 800 molecules per second (see~\autoref{tab:screening_throughput}). This allows screening of the entire ChEMBL library in under one hour per biological context on a single A100 GPU. In contrast, alternative methods such as DrugClip, Planet, Glide and even modern GPU-accelerated GNINA would require days to years. These results further position \modelname~as a scalable and efficient alternative for molecule prioritization.

\begin{table}[htb]
\centering
    \caption{Comparison of molecule prioritization performance on LIT-PCBA benchmark.}
    \label{tab:pcba-gpcr}
    \begin{threeparttable}  
    \begin{tabular}{lcc}
    \toprule
     {\textbf{Method}} & 
    {\textbf{ROC-AUC}} &
    {\textbf{EF@1\%}} \\
    \midrule
    2D ECFP4 search\tnote{a} & - & 2.49 \\ 
     FragSite (\footnotesize{\citeauthor{zhou2021fragsite}})\tnote{a} & - & 4.78 \\ 
    Glide-SP (\footnotesize{\citeauthor{halgren2004glide}})\tnote{b} & 0.532 & 3.41  \\ 
    Surflex (\footnotesize{\citeauthor{spitzer2012surflex}})\tnote{b}  & 0.515 & 2.50  \\
    $\Delta$vinaRF\textsubscript{20}(\footnotesize{\citeauthor{wang2017improving}})\tnote{a}  & - & 5.38  \\
    Gnina (\footnotesize{\citeauthor{mcnutt2021gnina}})\tnote{b} & \textbf{0.609} & 4.63  \\ 
    DeepDTA (\footnotesize{\citeauthor{ozturk2018deepdta}})\tnote{b} & 0.563 & 1.47  \\ 
    BigBind (\footnotesize{\citeauthor{brocidiacono2023bigbind}})\tnote{b} & 0.608 & 3.82  \\ 
    Planet (\footnotesize{\citeauthor{zhang2023planet}})\tnote{b} & 0.573 & 3.87  \\ 
    DrugCLIP (\footnotesize{\citeauthor{gao2024drugclip}})\tnote{b} & 0.572 & 5.51  \\ 
    \textbf{\modelname~(\footnotesize{Ours})} & 0.568 & \textbf{6.63}  \\ 
    \bottomrule
    \end{tabular}
    \begin{tablenotes}
        \item [a] as reported in \cite{zhou2021fragsite}
        \item [b] as reported in \cite{gao2024drugclip}
    \end{tablenotes}
    \end{threeparttable}
\end{table}

\subsection{Generalization to Unseen Targets.}
We further evaluated \modelname's ability to generalize to unseen protein targets---those completely absent from the training set and embedding space. This setting simulates practical scenarios where a new target of interest emerges after model deployment. We extended the embedding layer with newly added tokens and tested \modelname~both in a true zero-shot setting (no further updates) and after 50-shot fine-tuning using actives from ChEMBL35 and 5,000 randomly sampled decoys from LIT-PCBA.

\autoref{tab:zero_finetune} summarizes performance on four novel targets. Zero-shot generalization was variable: while \modelname~prioritized active compounds for SUMO E1/E2 and Plectin, performance was limited for BCDIN3D and SH2D1A. However, after just 50 active examples, the model achieved substantial improvements, reaching ROC-AUCs above 0.90 and enrichment factors exceeding 80 (EF@1\%) for SH2D1A. These results highlight the practicality of \modelname~for rapid adaptation to new targets with minimal data, offering a viable route for early-stage virtual screening.

\begin{table}[htb]
\centering
\caption{Performance on unseen protein targets: zero-shot vs. 50-shot fine-tuning. Decoys were sampled from LIT-PCBA.}
\label{tab:zero_finetune}
\resizebox{\linewidth}{!}{
\begin{tabular}{lcccccc}
\toprule
\multirow{2}{*}{\textbf{Target}} & \multicolumn{3}{c}{\textbf{0-shot}} & \multicolumn{3}{c}{\textbf{50-shot}} \\
\cmidrule(lr){2-4} \cmidrule(lr){5-7}
& ROC-AUC & EF@1\% & EF@5\% & ROC-AUC & EF@1\% & EF@5\% \\
\midrule
Plectin (202 actives)     & 0.672 & 5.47  & 4.44  & 0.689 & 5.58  & 4.46 \\
SUMO E1/E2 (83 actives)   & 0.782 & 14.51 & 4.11  & 0.823 & 44.42 & 12.39 \\
SH2D1A (59 actives)       & 0.740 & 0.00  & 0.00  & 0.998 & 85.83 & 20.03 \\
BCDIN3D RNA methyltransferase (70 actives)      & 0.401 & 0.00  & 0.00  & 0.909 & 19.03 & 9.22 \\
\bottomrule
\end{tabular}
}
\end{table}

\subsection{Activity Cliff Detection Performance}
\label{sec:ac_prediction}

Activity cliffs occur when minor structural modifications lead to large activity changes, making them critical for SAR analysis.  We evaluated \modelname~on activity cliff detection using the ACNet benchmark~\citep{zhang2023activity}. To ensure fair evaluation, we created new random splits that exclude compounds present in our training data. Following ACNet's protocol, we report ROC-AUC scores.

As shown in \autoref{tab:activity_cliff}, we benchmark against both traditional fingerprint-based approaches and graph neural networks. The pre-trained \modelname\textsubscript{pre} model performs near-randomly (ROC-AUC: 0.541), while biological context fine-tuning (\modelname) provides modest improvements (ROC-AUC: 0.581). However, preference-tuned \modelname\textsubscript{post} achieves performance (0.947) comparable to the best supervised baselines ECFP+MLP (0.946) and GRU (0.951). The substantial performance gain from preference tuning suggests that while \modelname~learns meaningful structure-activity patterns, the likelihood calibration provided by preference tuning is crucial for activity cliff detection, aligning with findings that activity cliffs often involve subtle changes~\citep{van2022exposing} that might not be captured by a conditional generation alone.

\begin{table}[hbt]
\caption{Performance comparison of \modelname~models in different model size on ACNet benchmark for activity cliff prediction}
\label{tab:activity_cliff}
\begin{center}
\small
\begin{tabular}{lc}
\toprule
         \textbf{Model }             & \textbf{ROC-AUC}  \\
\midrule
ECFP+MLP   & 0.946 $\pm$ 0.004 \\
GCN   & 0.918 $\pm$ 0.001 \\
GRU   & \textbf{0.951 $\pm$ 0.003}   \\ 
\modelname\textsubscript{pre}   & 0.541 $\pm$ 0.009  \\
\modelname~                     & 0.581 $\pm$ 0.003    \\
\modelname\textsubscript{post}  & 0.947 $\pm$ 0.000  \\
\bottomrule
\end{tabular}
\end{center}
\end{table}

\subsection{Fragment Attribution Reveals Structure-Activity Relationships}
\label{sec:sar_results}

We demonstrate \modelname's interpretability through fragment attribution analysis on a well-characterized drug. For each fragment $f_i$ in a molecule $x$, we compute its contribution $\psi_{f_i}$ by comparing the likelihood landscape when substituting $f_i$ with alternative fragments (see Appendix~\ref{appendix:sar-analysis}).

\begin{figure}[tb]
    \begin{center}
\centerline{\includegraphics[width=1.1\columnwidth]{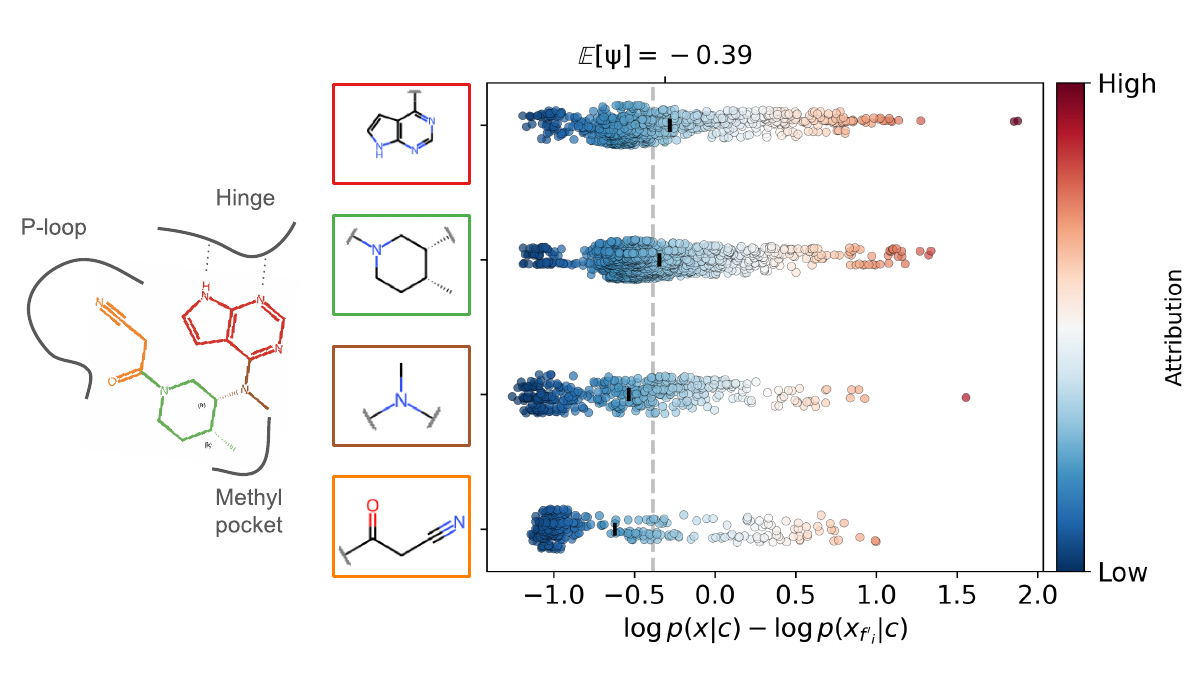}}
\caption{Fragment attribution analysis for Tofacitinib, a JAK inhibitor. Each point represents a molecule where a SAFE fragment replaces the one shown on the y-axis. Key pharmacophoric structures are highlighted: pyrrolopyrimidine hinge-binding core (red), methyl-substituted piperidine scaffold (green), and cyanoacetamide P-loop binding group (orange). Points above zero on the x-axis indicate replacements that decrease predicted likelihood; color intensity reflects attribution score.}
\label{fig:fragment_attribution}
\end{center}
\end{figure}

Figure~\ref{fig:fragment_attribution} shows \modelname's attribution scores for key fragments in Tofacitinib, an FDA-approved JAK inhibitor used for the treatment of rheumatoid arthritis, psoriatic arthritis, ulcerative colitis, and ankylosing spondylitis~\citep{jakclark2014discovery}. Each point represents a molecule with a single fragment replacement sampled by \modelname~or \modelname\textsubscript{pre}, with the x-axis showing normalized log-likelihood difference under JAK1 inhibition context.  The pyrrolopyrimidine core (red) shows the strongest attribution, reflecting its essential hinge-binding role. The methyl-substituted piperidine scaffold (green) shows moderate attribution, consistent with its contribution to kinome selectivity, while the cyanoacetamide group (orange) ranks lowest, consistent with its role in drug-like property optimization.

We conducted a similar analysis on Ruxolitinib, comparing \modelname~to GEAM~\citep{lee2023drug}, a pretrained molecular design model with fragment-level attributions. As shown in ~\autoref{fig:jak2_comparison_overall} of the Appendix, \modelname~again correctly recover key fragments, while GEAM assigns highest importance to the cyclopentyl group, a known non-essential fragment. These results, consistent with SAR studies \citep{jaksanachai2020insights,jakclark2014discovery,jaknoji2020discovery, zhou2014specificity,davis2021structural}, demonstrate \modelname’s potential for rational design and interpretable SAR exploration.

\subsection{Ablation Studies}
\label{sec:ablation}

We examine three key aspects of \modelname~through systematic ablation studies: the impact of model size and training stages, the effect of likelihood normalization strategies, and the role of biological context modeling. Detailed analyses are provided in Appendix~\ref{appendix:ablation}.

\paragraph{Model Size and Training Stages} : We experiment with three model sizes ($\sim$10M, $\sim$25M and $\sim$45M parameters) across multiple tasks. For molecular generation, model size has minimal impact on core metrics, with all variants maintaining high validity ($>$99\%) and diversity ($\sim$85\%). However, preference tuning (\modelname\textsubscript{post}) significantly impacts generation capability, decreasing validity to $\sim$40\% on most tasks. This degradation can be partially mitigated by larger models, as the $\sim$45M parameter model can maintain validity above 95\%.

For predictive tasks, model scaling consistently improves performance, with larger models showing 5-10\% gains in ROC-AUC scores for target prediction. Preference tuning shows task-dependent effects: while it slightly reduces absolute performance metrics in molecular prioritization and DTI prediction, it dramatically improves activity cliff prediction (ROC-AUC from $\sim$0.57 to $\sim$0.95). Notably, models without preference tuning perform no better than random, highlighting that explicit optimization for structure-activity relationships is essential for capturing subtle chemical changes that lead to large activity differences.
 
\paragraph{Likelihood Normalization Strategies}: We evaluate multiple strategies for normalizing likelihood scores from our conditional model: population-based normalization ($l_\text{pop}$), which contrasts molecules against a reference population sampled under the same biological context; null-condition normalization ($l_\text{null}$), which measures information gain relative to an unconditional baseline; and their prior-adjusted variants ($l_\text{prior}$) (see Appendix~\ref{appendix:lkl_normalization}). Our empirical evaluation reveals task-specific effectiveness: $l_\text{null}$ excels in classification tasks where discriminating between biological contexts is paramount, while $l_\text{pop}$ proves superior for in-context molecule prioritization, particularly in out-of-distribution settings where calibrated within-context comparisons are crucial. Prior adjustment ($l\text{prior}$) did not provide any clear advantage, though in our case, we use a frequentist perspective based on MoAT-DB target representation statistics.

\paragraph{Impact of Biological Context Modeling} :  Analysis of \modelname's predictive capabilities reveals the role of mechanism of action (MoA) information. Removing MoA context in DTI classification typically leads to performance drops in IID settings. This degradation is most pronounced for targets with multiple known mechanisms, where MoA information helps disambiguate different binding modes. However, as such information is not always readily available, masking out MoA remains competitive for DTI prediction.

\section{Conclusion}
\label{sec:conclusion}
We presented \modelname, a conditional chemical language model that integrates fragment-based representations with biological context for targeted molecular design. Through comprehensive evaluations, we demonstrated its effectiveness in zero-shot drug discovery tasks, from molecule prioritization to SAR analysis. While \modelname~mitigates the lack of explicit reward signals, it complements rather than replaces traditional optimization when reliable activity data is available. Its masking strategy ensures robustness to partial or missing biological context, though performance depends on annotation quality.

Despite performance gains with model scaling, data availability remains a key bottleneck. MoAT-DB’s focus on active compounds, including weak binders, also limits exposure to inactive chemical space, which is crucial for tasks like activity cliff prediction. 

Future work could enhance \modelname~with structural data and explicit polypharmacology modeling. Its fragment attribution framework also enables mechanistic interpretation of drug-target interactions, offering a promising direction for integrating chemical and biological knowledge in early-stage drug discovery.

\section*{Impact Statement}
Our work advances conditional molecular generation for drug discovery applications. \modelname~could significantly reduce the costs and timeframes of early-stage drug discovery by enabling more focused screening campaigns and providing interpretable structure-activity insights for lead optimization, potentially leading to safer and more effective therapeutics. While these capabilities could potentially be misused to generate harmful substances, we mitigate this risk by focusing on drug-like chemical space and therapeutic contexts. The framework's fragment attribution analysis offers medicinal chemists actionable insights for compound design, potentially reducing experimental iterations needed. As model scaling shows clear benefits, we acknowledge the environmental impact of large-scale training and emphasize the importance of data quality and efficient architectures.

\bibliography{references}  

\begin{thebibliography}{94}
\providecommand{\natexlab}[1]{#1}
\providecommand{\url}[1]{\texttt{#1}}
\expandafter\ifx\csname urlstyle\endcsname\relax
  \providecommand{\doi}[1]{doi: #1}\else
  \providecommand{\doi}{doi: \begingroup \urlstyle{rm}\Url}\fi

\bibitem[Brocidiacono et~al.(2023)Brocidiacono, Francoeur, Aggarwal, Popov, Koes, and Tropsha]{brocidiacono2023bigbind}
Brocidiacono, M., Francoeur, P., Aggarwal, R., Popov, K.~I., Koes, D.~R., and Tropsha, A.
\newblock Bigbind: learning from nonstructural data for structure-based virtual screening.
\newblock \emph{Journal of Chemical Information and Modeling}, 64\penalty0 (7):\penalty0 2488--2495, 2023.

\bibitem[Clark et~al.(2014)Clark, Flanagan, and Telliez]{jakclark2014discovery}
Clark, J.~D., Flanagan, M.~E., and Telliez, J.-B.
\newblock Discovery and development of janus kinase (jak) inhibitors for inflammatory diseases: Miniperspective.
\newblock \emph{Journal of medicinal chemistry}, 57\penalty0 (12):\penalty0 5023--5038, 2014.

\bibitem[Copeland et~al.(2006)Copeland, Pompliano, and Meek]{copeland2006drug}
Copeland, R.~A., Pompliano, D.~L., and Meek, T.~D.
\newblock Drug--target residence time and its implications for lead optimization.
\newblock \emph{Nature reviews Drug discovery}, 5\penalty0 (9):\penalty0 730--739, 2006.

\bibitem[Davis et~al.(2011)Davis, Hunt, Herrgard, Ciceri, Wodicka, Pallares, Hocker, Treiber, and Zarrinkar]{davis2011comprehensive}
Davis, M.~I., Hunt, J.~P., Herrgard, S., Ciceri, P., Wodicka, L.~M., Pallares, G., Hocker, M., Treiber, D.~K., and Zarrinkar, P.~P.
\newblock Comprehensive analysis of kinase inhibitor selectivity.
\newblock \emph{Nature biotechnology}, 29\penalty0 (11):\penalty0 1046--1051, 2011.

\bibitem[Davis(2020)]{davis2020mechanism}
Davis, R.~L.
\newblock Mechanism of action and target identification: a matter of timing in drug discovery.
\newblock \emph{Iscience}, 23\penalty0 (9), 2020.

\bibitem[Davis et~al.(2021)Davis, Li, Yun, Chan, Nareddy, Gunawan, Ayaz, Lawrence, Reuther, Lawrence, et~al.]{davis2021structural}
Davis, R.~R., Li, B., Yun, S.~Y., Chan, A., Nareddy, P., Gunawan, S., Ayaz, M., Lawrence, H.~R., Reuther, G.~W., Lawrence, N.~J., et~al.
\newblock Structural insights into jak2 inhibition by ruxolitinib, fedratinib, and derivatives thereof.
\newblock \emph{Journal of medicinal chemistry}, 64\penalty0 (4):\penalty0 2228--2241, 2021.

\bibitem[Degen et~al.(2008)Degen, Wegscheid-Gerlach, Zaliani, and Rarey]{degen2008art}
Degen, J., Wegscheid-Gerlach, C., Zaliani, A., and Rarey, M.
\newblock On the art of compiling and using'drug-like'chemical fragment spaces.
\newblock \emph{ChemMedChem}, 3\penalty0 (10):\penalty0 1503, 2008.

\bibitem[Dobberstein et~al.(2024)Dobberstein, Maass, and Hamaekers]{dobberstein2024llamol}
Dobberstein, N., Maass, A., and Hamaekers, J.
\newblock Llamol: a dynamic multi-conditional generative transformer for de novo molecular design.
\newblock \emph{Journal of Cheminformatics}, 16\penalty0 (1):\penalty0 73, 2024.

\bibitem[Ewing et~al.(2001)Ewing, Makino, Skillman, and Kuntz]{ewing2001dock}
Ewing, T.~J., Makino, S., Skillman, A.~G., and Kuntz, I.~D.
\newblock Dock 4.0: search strategies for automated molecular docking of flexible molecule databases.
\newblock \emph{Journal of computer-aided molecular design}, 15:\penalty0 411--428, 2001.

\bibitem[Feng et~al.(2024)Feng, Wang, Lin, Zhu, Wang, Dong, Bai, Wang, Zhou, Peng, et~al.]{feng2024generation}
Feng, W., Wang, L., Lin, Z., Zhu, Y., Wang, H., Dong, J., Bai, R., Wang, H., Zhou, J., Peng, W., et~al.
\newblock Generation of 3d molecules in pockets via a language model.
\newblock \emph{Nature Machine Intelligence}, 6\penalty0 (1):\penalty0 62--73, 2024.

\bibitem[Gao et~al.(2024)Gao, Qiang, Tan, Jia, Ren, Lu, Liu, Ma, and Lan]{gao2024drugclip}
Gao, B., Qiang, B., Tan, H., Jia, Y., Ren, M., Lu, M., Liu, J., Ma, W.-Y., and Lan, Y.
\newblock Drugclip: Contrasive protein-molecule representation learning for virtual screening.
\newblock \emph{Advances in Neural Information Processing Systems}, 36, 2024.

\bibitem[Gao et~al.(2022)Gao, Fu, Sun, and Coley]{Gao2022}
Gao, W., Fu, T., Sun, J., and Coley, C.~W.
\newblock Sample efficiency matters: A benchmark for practical molecular optimization.
\newblock \emph{Advances in Neural Information Processing Systems}, 35, 6 2022.
\newblock ISSN 10495258.
\newblock URL \url{https://arxiv.org/abs/2206.12411v2}.

\bibitem[G{\'o}mez-Bombarelli et~al.(2018)G{\'o}mez-Bombarelli, Wei, Duvenaud, Hern{\'a}ndez-Lobato, S{\'a}nchez-Lengeling, Sheberla, Aguilera-Iparraguirre, Hirzel, Adams, and Aspuru-Guzik]{gomez2018automatic}
G{\'o}mez-Bombarelli, R., Wei, J.~N., Duvenaud, D., Hern{\'a}ndez-Lobato, J.~M., S{\'a}nchez-Lengeling, B., Sheberla, D., Aguilera-Iparraguirre, J., Hirzel, T.~D., Adams, R.~P., and Aspuru-Guzik, A.
\newblock Automatic chemical design using a data-driven continuous representation of molecules.
\newblock \emph{ACS central science}, 4\penalty0 (2):\penalty0 268--276, 2018.

\bibitem[Grimm et~al.(2020)Grimm, Liu, Yang, Bu, Xiao, and Cao]{Grimm2020}
Grimm, M., Liu, Y., Yang, X., Bu, C., Xiao, Z., and Cao, Y.
\newblock Ligmate: a multifeature integration algorithm for ligand-similarity-based virtual screening.
\newblock \emph{Journal of Chemical Information and Modeling}, 60\penalty0 (12):\penalty0 6044--6053, 2020.

\bibitem[Grisoni(2023)]{grisoni2023chemical}
Grisoni, F.
\newblock Chemical language models for de novo drug design: Challenges and opportunities.
\newblock \emph{Current Opinion in Structural Biology}, 79:\penalty0 102527, 2023.

\bibitem[Halgren et~al.(2004)Halgren, Murphy, Friesner, Beard, Frye, Pollard, and Banks]{halgren2004glide}
Halgren, T.~A., Murphy, R.~B., Friesner, R.~A., Beard, H.~S., Frye, L.~L., Pollard, W.~T., and Banks, J.~L.
\newblock Glide: a new approach for rapid, accurate docking and scoring. 2. enrichment factors in database screening.
\newblock \emph{Journal of medicinal chemistry}, 47\penalty0 (7):\penalty0 1750--1759, 2004.

\bibitem[H{\"o}dl et~al.(2023)H{\"o}dl, Robinson, Bachrach, Huck, and Kachman]{hodl2023explainability}
H{\"o}dl, S., Robinson, W., Bachrach, Y., Huck, W., and Kachman, T.
\newblock Explainability techniques for chemical language models.
\newblock \emph{arXiv preprint arXiv:2305.16192}, 2023.

\bibitem[Hussain \& Rea(2010)Hussain and Rea]{hussain2010computationally}
Hussain, J. and Rea, C.
\newblock Computationally efficient algorithm to identify matched molecular pairs (mmps) in large data sets.
\newblock \emph{Journal of chemical information and modeling}, 50\penalty0 (3):\penalty0 339--348, 2010.

\bibitem[Imrie et~al.(2021)Imrie, Hadfield, Bradley, and Deane]{imrie2021deep}
Imrie, F., Hadfield, T.~E., Bradley, A.~R., and Deane, C.~M.
\newblock Deep generative design with 3d pharmacophoric constraints.
\newblock \emph{Chemical science}, 12\penalty0 (43):\penalty0 14577--14589, 2021.

\bibitem[Irwin et~al.(2020)Irwin, Tang, Young, Dandarchuluun, Wong, Khurelbaatar, Moroz, Mayfield, and Sayle]{irwin2020zinc20}
Irwin, J.~J., Tang, K.~G., Young, J., Dandarchuluun, C., Wong, B.~R., Khurelbaatar, M., Moroz, Y.~S., Mayfield, J., and Sayle, R.~A.
\newblock Zinc20—a free ultralarge-scale chemical database for ligand discovery.
\newblock \emph{Journal of chemical information and modeling}, 60\penalty0 (12):\penalty0 6065--6073, 2020.

\bibitem[Irwin et~al.(2022)Irwin, Dimitriadis, He, and Bjerrum]{irwin2022chemformer}
Irwin, R., Dimitriadis, S., He, J., and Bjerrum, E.~J.
\newblock Chemformer: a pre-trained transformer for computational chemistry.
\newblock \emph{Machine Learning: Science and Technology}, 3\penalty0 (1):\penalty0 015022, 2022.

\bibitem[Jhoti et~al.(2013)Jhoti, Williams, Rees, and Murray]{Jhoti2013}
Jhoti, H., Williams, G., Rees, D.~C., and Murray, C.~W.
\newblock The 'rule of three' for fragment-based drug discovery: where are we now?
\newblock \emph{Nature Reviews Drug Discovery 2013 12:8}, 12:\penalty0 644--644, 7 2013.
\newblock ISSN 1474-1784.
\newblock \doi{10.1038/nrd3926-c1}.

\bibitem[Jiang et~al.(2024)Jiang, Zhang, You, Zhang, Yao, Xie, Zhang, Xia, Dai, Wu, et~al.]{jiang2024pocketflow}
Jiang, Y., Zhang, G., You, J., Zhang, H., Yao, R., Xie, H., Zhang, L., Xia, Z., Dai, M., Wu, Y., et~al.
\newblock Pocketflow is a data-and-knowledge-driven structure-based molecular generative model.
\newblock \emph{Nature Machine Intelligence}, 6\penalty0 (3):\penalty0 326--337, 2024.

\bibitem[Jim{\'e}nez-Luna et~al.(2020)Jim{\'e}nez-Luna, Grisoni, and Schneider]{jimenez2020drug}
Jim{\'e}nez-Luna, J., Grisoni, F., and Schneider, G.
\newblock Drug discovery with explainable artificial intelligence.
\newblock \emph{Nature Machine Intelligence}, 2\penalty0 (10):\penalty0 573--584, 2020.

\bibitem[Jin et~al.(2020)Jin, Barzilay, and Jaakkola]{jin2020hierarchical}
Jin, W., Barzilay, R., and Jaakkola, T.
\newblock Hierarchical generation of molecular graphs using structural motifs.
\newblock In \emph{International conference on machine learning}, pp.\  4839--4848. PMLR, 2020.

\bibitem[Kang \& Cho(2018)Kang and Cho]{kang2018conditional}
Kang, S. and Cho, K.
\newblock Conditional molecular design with deep generative models.
\newblock \emph{Journal of chemical information and modeling}, 59\penalty0 (1):\penalty0 43--52, 2018.

\bibitem[Kenakin(2019)]{kenakin2019biased}
Kenakin, T.
\newblock Biased receptor signaling in drug discovery.
\newblock \emph{Pharmacological reviews}, 71\penalty0 (2):\penalty0 267--315, 2019.

\bibitem[Kotsias et~al.(2020)Kotsias, Ar{\'u}s-Pous, Chen, Engkvist, Tyrchan, and Bjerrum]{kotsias2020direct}
Kotsias, P.-C., Ar{\'u}s-Pous, J., Chen, H., Engkvist, O., Tyrchan, C., and Bjerrum, E.~J.
\newblock Direct steering of de novo molecular generation with descriptor conditional recurrent neural networks.
\newblock \emph{Nature Machine Intelligence}, 2\penalty0 (5):\penalty0 254--265, 2020.

\bibitem[Lee et~al.(2023)Lee, Lee, Kawaguchi, and Hwang]{lee2023drug}
Lee, S., Lee, S., Kawaguchi, K., and Hwang, S.~J.
\newblock Drug discovery with dynamic goal-aware fragments.
\newblock \emph{arXiv preprint arXiv:2310.00841}, 2023.

\bibitem[Lee et~al.(2025)Lee, Kreis, Veccham, Liu, Reidenbach, Peng, Paliwal, Nie, and Vahdat]{lee2025genmol}
Lee, S., Kreis, K., Veccham, S.~P., Liu, M., Reidenbach, D., Peng, Y., Paliwal, S., Nie, W., and Vahdat, A.
\newblock Genmol: A drug discovery generalist with discrete diffusion.
\newblock \emph{arXiv preprint arXiv:2501.06158}, 2025.

\bibitem[Li et~al.(2018)Li, Zhang, and Liu]{li2018multi}
Li, Y., Zhang, L., and Liu, Z.
\newblock Multi-objective de novo drug design with conditional graph generative model.
\newblock \emph{Journal of cheminformatics}, 10:\penalty0 1--24, 2018.

\bibitem[Lin et~al.(2019)Lin, Giuliano, Palladino, John, Abramowicz, Yuan, Sausville, Lukow, Liu, Chait, et~al.]{lin2019off}
Lin, A., Giuliano, C.~J., Palladino, A., John, K.~M., Abramowicz, C., Yuan, M.~L., Sausville, E.~L., Lukow, D.~A., Liu, L., Chait, A.~R., et~al.
\newblock Off-target toxicity is a common mechanism of action of cancer drugs undergoing clinical trials.
\newblock \emph{Science translational medicine}, 11\penalty0 (509):\penalty0 eaaw8412, 2019.

\bibitem[Lin et~al.(2024)Lin, Huang, Zhang, Liu, Wu, Li, Chen, and Li]{lin2024functional}
Lin, H., Huang, Y., Zhang, O., Liu, Y., Wu, L., Li, S., Chen, Z., and Li, S.~Z.
\newblock Functional-group-based diffusion for pocket-specific molecule generation and elaboration.
\newblock \emph{Advances in Neural Information Processing Systems}, 36, 2024.

\bibitem[Liu et~al.(2022)Liu, Luo, Uchino, Maruhashi, and Ji]{liu2022generating}
Liu, M., Luo, Y., Uchino, K., Maruhashi, K., and Ji, S.
\newblock Generating 3d molecules for target protein binding.
\newblock \emph{arXiv preprint arXiv:2204.09410}, 2022.

\bibitem[Loeffler et~al.(2024)Loeffler, He, Tibo, Janet, Voronov, Mervin, and Engkvist]{loeffler2024reinvent}
Loeffler, H.~H., He, J., Tibo, A., Janet, J.~P., Voronov, A., Mervin, L.~H., and Engkvist, O.
\newblock Reinvent 4: Modern ai--driven generative molecule design.
\newblock \emph{Journal of Cheminformatics}, 16\penalty0 (1):\penalty0 20, 2024.

\bibitem[Lundberg(2017)]{lundberg2017unified}
Lundberg, S.
\newblock A unified approach to interpreting model predictions.
\newblock \emph{arXiv preprint arXiv:1705.07874}, 2017.

\bibitem[Lyu et~al.(2019)Lyu, Wang, Balius, Singh, Levit, Moroz, O’Meara, Che, Algaa, Tolmachova, et~al.]{lyu2019ultra}
Lyu, J., Wang, S., Balius, T.~E., Singh, I., Levit, A., Moroz, Y.~S., O’Meara, M.~J., Che, T., Algaa, E., Tolmachova, K., et~al.
\newblock Ultra-large library docking for discovering new chemotypes.
\newblock \emph{Nature}, 566\penalty0 (7743):\penalty0 224--229, 2019.

\bibitem[Mastrolorito et~al.(2024)Mastrolorito, Ciriaco, Togo, Gambacorta, Trisciuzzi, Altomare, Amoroso, Grisoni, and Nicolotti]{mastrolorito2024fragsmiles}
Mastrolorito, F., Ciriaco, F., Togo, M.~V., Gambacorta, N., Trisciuzzi, D., Altomare, C.~D., Amoroso, N., Grisoni, F., and Nicolotti, O.
\newblock fragsmiles: A chemical string notation for advanced fragment and chirality representation.
\newblock 2024.

\bibitem[McNutt et~al.(2021)McNutt, Francoeur, Aggarwal, Masuda, Meli, Ragoza, Sunseri, and Koes]{mcnutt2021gnina}
McNutt, A.~T., Francoeur, P., Aggarwal, R., Masuda, T., Meli, R., Ragoza, M., Sunseri, J., and Koes, D.~R.
\newblock Gnina 1.0: molecular docking with deep learning.
\newblock \emph{Journal of cheminformatics}, 13\penalty0 (1):\penalty0 43, 2021.

\bibitem[M{\'e}ndez-Lucio et~al.(2020)M{\'e}ndez-Lucio, Baillif, Clevert, Rouqui{\'e}, and Wichard]{mendez2020novo}
M{\'e}ndez-Lucio, O., Baillif, B., Clevert, D.-A., Rouqui{\'e}, D., and Wichard, J.
\newblock De novo generation of hit-like molecules from gene expression signatures using artificial intelligence.
\newblock \emph{Nature communications}, 11\penalty0 (1):\penalty0 10, 2020.

\bibitem[Mesbahi \& Noutahi(2024)Mesbahi and Noutahi]{mesbahi2024safe}
Mesbahi, Y.~E. and Noutahi, E.
\newblock Safe setup for generative molecular design.
\newblock \emph{arXiv preprint arXiv:2410.20232}, 2024.

\bibitem[Moret et~al.(2023)Moret, Pachon~Angona, Cotos, Yan, Atz, Brunner, Baumgartner, Grisoni, and Schneider]{moret2023leveraging}
Moret, M., Pachon~Angona, I., Cotos, L., Yan, S., Atz, K., Brunner, C., Baumgartner, M., Grisoni, F., and Schneider, G.
\newblock Leveraging molecular structure and bioactivity with chemical language models for de novo drug design.
\newblock \emph{Nature Communications}, 14\penalty0 (1):\penalty0 114, 2023.

\bibitem[Mysinger et~al.(2012)Mysinger, Carchia, Irwin, and Shoichet]{mysinger2012dude}
Mysinger, M.~M., Carchia, M., Irwin, J.~J., and Shoichet, B.~K.
\newblock Directory of useful decoys, enhanced (dud-e): better ligands and decoys for better benchmarking.
\newblock \emph{Journal of medicinal chemistry}, 55\penalty0 (14):\penalty0 6582--6594, 2012.

\bibitem[Neudert \& Klebe(2011)Neudert and Klebe]{neudert2011dsx}
Neudert, G. and Klebe, G.
\newblock Dsx: a knowledge-based scoring function for the assessment of protein--ligand complexes.
\newblock \emph{Journal of chemical information and modeling}, 51\penalty0 (10):\penalty0 2731--2745, 2011.

\bibitem[Noji et~al.(2020)Noji, Hara, Miura, Yamanaka, Maeda, Hori, Yamamoto, Obika, Inoue, Hase, et~al.]{jaknoji2020discovery}
Noji, S., Hara, Y., Miura, T., Yamanaka, H., Maeda, K., Hori, A., Yamamoto, H., Obika, S., Inoue, M., Hase, Y., et~al.
\newblock Discovery of a janus kinase inhibitor bearing a highly three-dimensional spiro scaffold: Jte-052 (delgocitinib) as a new dermatological agent to treat inflammatory skin disorders.
\newblock \emph{Journal of Medicinal Chemistry}, 63\penalty0 (13):\penalty0 7163--7185, 2020.

\bibitem[Noutahi et~al.(2024)Noutahi, Gabellini, Craig, Lim, and Tossou]{noutahi2024gotta}
Noutahi, E., Gabellini, C., Craig, M., Lim, J.~S., and Tossou, P.
\newblock Gotta be safe: a new framework for molecular design.
\newblock \emph{Digital Discovery}, 3\penalty0 (4):\penalty0 796--804, 2024.

\bibitem[Olivecrona et~al.(2017)Olivecrona, Blaschke, Engkvist, and Chen]{olivecrona2017molecular}
Olivecrona, M., Blaschke, T., Engkvist, O., and Chen, H.
\newblock Molecular de-novo design through deep reinforcement learning.
\newblock \emph{Journal of cheminformatics}, 9:\penalty0 1--14, 2017.

\bibitem[{\"O}zt{\"u}rk et~al.(2018){\"O}zt{\"u}rk, {\"O}zg{\"u}r, and Ozkirimli]{ozturk2018deepdta}
{\"O}zt{\"u}rk, H., {\"O}zg{\"u}r, A., and Ozkirimli, E.
\newblock Deepdta: deep drug--target binding affinity prediction.
\newblock \emph{Bioinformatics}, 34\penalty0 (17):\penalty0 i821--i829, 2018.

\bibitem[Paggi et~al.(2024)Paggi, Pandit, and Dror]{paggi2024art}
Paggi, J.~M., Pandit, A., and Dror, R.~O.
\newblock The art and science of molecular docking.
\newblock \emph{Annual Review of Biochemistry}, 93, 2024.

\bibitem[Pham et~al.(2022)Pham, Xie, and Zhang]{pham2022fame}
Pham, T.-H., Xie, L., and Zhang, P.
\newblock Fame: fragment-based conditional molecular generation for phenotypic drug discovery.
\newblock In \emph{Proceedings of the 2022 SIAM International Conference on Data Mining (SDM)}, pp.\  720--728. SIAM, 2022.

\bibitem[Qian et~al.(2024)Qian, Huang, Tu, and Xu]{qian2024kgdiff}
Qian, H., Huang, W., Tu, S., and Xu, L.
\newblock Kgdiff: towards explainable target-aware molecule generation with knowledge guidance.
\newblock \emph{Briefings in Bioinformatics}, 25\penalty0 (1):\penalty0 bbad435, 2024.

\bibitem[Ragoza et~al.(2022)Ragoza, Masuda, and Koes]{ragoza2022generating}
Ragoza, M., Masuda, T., and Koes, D.~R.
\newblock Generating 3d molecules conditional on receptor binding sites with deep generative models.
\newblock \emph{Chemical science}, 13\penalty0 (9):\penalty0 2701--2713, 2022.

\bibitem[Ribeiro et~al.(2016)Ribeiro, Singh, and Guestrin]{ribeiro2016should}
Ribeiro, M.~T., Singh, S., and Guestrin, C.
\newblock " why should i trust you?" explaining the predictions of any classifier.
\newblock In \emph{Proceedings of the 22nd ACM SIGKDD international conference on knowledge discovery and data mining}, pp.\  1135--1144, 2016.

\bibitem[Sanachai et~al.(2020)Sanachai, Mahalapbutr, Choowongkomon, Poo-Arporn, Wolschann, and Rungrotmongkol]{jaksanachai2020insights}
Sanachai, K., Mahalapbutr, P., Choowongkomon, K., Poo-Arporn, R.~P., Wolschann, P., and Rungrotmongkol, T.
\newblock Insights into the binding recognition and susceptibility of tofacitinib toward janus kinases.
\newblock \emph{ACS omega}, 5\penalty0 (1):\penalty0 369--377, 2020.

\bibitem[Santos-Martins et~al.(2021)Santos-Martins, Solis-Vasquez, Tillack, Sanner, Koch, and Forli]{santos2021autodock}
Santos-Martins, D., Solis-Vasquez, L., Tillack, A.~F., Sanner, M.~F., Koch, A., and Forli, S.
\newblock Accelerating autodock4 with gpus and gradient-based local search.
\newblock \emph{Journal of chemical theory and computation}, 17\penalty0 (2):\penalty0 1060--1073, 2021.

\bibitem[Sarti et~al.(2023)Sarti, Feldhus, Sickert, Van Der~Wal, Nissim, and Bisazza]{sarti2023inseq}
Sarti, G., Feldhus, N., Sickert, L., Van Der~Wal, O., Nissim, M., and Bisazza, A.
\newblock Inseq: An interpretability toolkit for sequence generation models.
\newblock \emph{arXiv preprint arXiv:2302.13942}, 2023.

\bibitem[Segler et~al.(2018)Segler, Kogej, Tyrchan, and Waller]{segler2018generating}
Segler, M.~H., Kogej, T., Tyrchan, C., and Waller, M.~P.
\newblock Generating focused molecule libraries for drug discovery with recurrent neural networks.
\newblock \emph{ACS central science}, 4\penalty0 (1):\penalty0 120--131, 2018.

\bibitem[Shapley(1953)]{shapley1953value}
Shapley, L.~S.
\newblock A value for n-person games.
\newblock \emph{Contribution to the Theory of Games}, 2, 1953.

\bibitem[Shen et~al.(2023)Shen, Zhao, Wu, Sun, Li, Yan, and Shao]{shen2023allosteric}
Shen, S., Zhao, C., Wu, C., Sun, S., Li, Z., Yan, W., and Shao, Z.
\newblock Allosteric modulation of g protein-coupled receptor signaling.
\newblock \emph{Frontiers in Endocrinology}, 14:\penalty0 1137604, 2023.

\bibitem[Shi et~al.(2024)Shi, Yang, Xie, Yin, and Zhang]{shi2024review}
Shi, W., Yang, H., Xie, L., Yin, X.-X., and Zhang, Y.
\newblock A review of machine learning-based methods for predicting drug--target interactions.
\newblock \emph{Health Information Science and Systems}, 12\penalty0 (1):\penalty0 1--16, 2024.

\bibitem[Skalic et~al.(2019)Skalic, Jim{\'e}nez, Sabbadin, and De~Fabritiis]{skalic2019shape}
Skalic, M., Jim{\'e}nez, J., Sabbadin, D., and De~Fabritiis, G.
\newblock Shape-based generative modeling for de novo drug design.
\newblock \emph{Journal of chemical information and modeling}, 59\penalty0 (3):\penalty0 1205--1214, 2019.

\bibitem[Spitzer \& Jain(2012)Spitzer and Jain]{spitzer2012surflex}
Spitzer, R. and Jain, A.~N.
\newblock Surflex-dock: Docking benchmarks and real-world application.
\newblock \emph{Journal of computer-aided molecular design}, 26:\penalty0 687--699, 2012.

\bibitem[Subramanian et~al.(2017)Subramanian, Narayan, Corsello, Peck, Natoli, Lu, Gould, Davis, Tubelli, Asiedu, et~al.]{subramanian2017next}
Subramanian, A., Narayan, R., Corsello, S.~M., Peck, D.~D., Natoli, T.~E., Lu, X., Gould, J., Davis, J.~F., Tubelli, A.~A., Asiedu, J.~K., et~al.
\newblock A next generation connectivity map: L1000 platform and the first 1,000,000 profiles.
\newblock \emph{Cell}, 171\penalty0 (6):\penalty0 1437--1452, 2017.

\bibitem[Sunseri \& Koes(2021)Sunseri and Koes]{sunseri2021virtual}
Sunseri, J. and Koes, D.~R.
\newblock Virtual screening with gnina 1.0.
\newblock \emph{Molecules}, 26\penalty0 (23):\penalty0 7369, 2021.

\bibitem[Swinney \& Anthony(2011)Swinney and Anthony]{swinney2011were}
Swinney, D.~C. and Anthony, J.
\newblock How were new medicines discovered?
\newblock \emph{Nature reviews Drug discovery}, 10\penalty0 (7):\penalty0 507--519, 2011.

\bibitem[Sypetkowski et~al.(2024)Sypetkowski, Wenkel, Poursafaei, Dickson, Labs, Fradkin, and Beaini]{Sypetkowski2024}
Sypetkowski, M., Wenkel, F., Poursafaei, F., Dickson, N., Labs, K. S.~V., Fradkin, M.~P., and Beaini, D.
\newblock On the scalability of gnns for molecular graphs.
\newblock 4 2024.

\bibitem[Tang et~al.(2014)Tang, Szwajda, Shakyawar, Xu, Hintsanen, Wennerberg, and Aittokallio]{tang2014making}
Tang, J., Szwajda, A., Shakyawar, S., Xu, T., Hintsanen, P., Wennerberg, K., and Aittokallio, T.
\newblock Making sense of large-scale kinase inhibitor bioactivity data sets: a comparative and integrative analysis.
\newblock \emph{Journal of Chemical Information and Modeling}, 54\penalty0 (3):\penalty0 735--743, 2014.

\bibitem[Tossou et~al.(2024)Tossou, Wognum, Craig, Mary, and Noutahi]{tossou2024real}
Tossou, P., Wognum, C., Craig, M., Mary, H., and Noutahi, E.
\newblock Real-world molecular out-of-distribution: Specification and investigation.
\newblock \emph{Journal of Chemical Information and Modeling}, 64\penalty0 (3):\penalty0 697--711, 2024.

\bibitem[Tran-Nguyen et~al.(2020)Tran-Nguyen, Jacquemard, and Rognan]{tran2020lit}
Tran-Nguyen, V.-K., Jacquemard, C., and Rognan, D.
\newblock Lit-pcba: an unbiased data set for machine learning and virtual screening.
\newblock \emph{Journal of chemical information and modeling}, 60\penalty0 (9):\penalty0 4263--4273, 2020.

\bibitem[Trott \& Olson(2010)Trott and Olson]{trott2010autodockvina}
Trott, O. and Olson, A.~J.
\newblock Autodock vina: improving the speed and accuracy of docking with a new scoring function, efficient optimization, and multithreading.
\newblock \emph{Journal of computational chemistry}, 31\penalty0 (2):\penalty0 455--461, 2010.

\bibitem[Truchon \& Bayly(2007)Truchon and Bayly]{truchon2007evaluating}
Truchon, J.-F. and Bayly, C.~I.
\newblock Evaluating virtual screening methods: good and bad metrics for the “early recognition” problem.
\newblock \emph{Journal of chemical information and modeling}, 47\penalty0 (2):\penalty0 488--508, 2007.

\bibitem[Van~Tilborg et~al.(2022)Van~Tilborg, Alenicheva, and Grisoni]{van2022exposing}
Van~Tilborg, D., Alenicheva, A., and Grisoni, F.
\newblock Exposing the limitations of molecular machine learning with activity cliffs.
\newblock \emph{Journal of chemical information and modeling}, 62\penalty0 (23):\penalty0 5938--5951, 2022.

\bibitem[Vig \& Belinkov(2019)Vig and Belinkov]{vig2019analyzing}
Vig, J. and Belinkov, Y.
\newblock Analyzing the structure of attention in a transformer language model.
\newblock \emph{arXiv preprint arXiv:1906.04284}, 2019.

\bibitem[Wachter et~al.(2017)Wachter, Mittelstadt, and Russell]{wachter2017counterfactual}
Wachter, S., Mittelstadt, B., and Russell, C.
\newblock Counterfactual explanations without opening the black box: Automated decisions and the gdpr.
\newblock \emph{Harv. JL \& Tech.}, 31:\penalty0 841, 2017.

\bibitem[Wacker et~al.(2017)Wacker, Wang, McCorvy, Betz, Venkatakrishnan, Levit, Lansu, Schools, Che, Nichols, et~al.]{wacker2017crystal}
Wacker, D., Wang, S., McCorvy, J.~D., Betz, R.~M., Venkatakrishnan, A., Levit, A., Lansu, K., Schools, Z.~L., Che, T., Nichols, D.~E., et~al.
\newblock Crystal structure of an lsd-bound human serotonin receptor.
\newblock \emph{Cell}, 168\penalty0 (3):\penalty0 377--389, 2017.

\bibitem[Wang \& Zhang(2017)Wang and Zhang]{wang2017improving}
Wang, C. and Zhang, Y.
\newblock Improving scoring-docking-screening powers of protein--ligand scoring functions using random forest.
\newblock \emph{Journal of computational chemistry}, 38\penalty0 (3):\penalty0 169--177, 2017.

\bibitem[Wang et~al.(2025)Wang, Schmidgall, Jaeger, Zhang, Pilgrim, Matias, Barral, Fleet, and Azizi]{wang2025txgemma}
Wang, E., Schmidgall, S., Jaeger, P.~F., Zhang, F., Pilgrim, R., Matias, Y., Barral, J., Fleet, D., and Azizi, S.
\newblock Txgemma: Efficient and agentic llms for therapeutics.
\newblock \emph{arXiv preprint arXiv:2504.06196}, 2025.

\bibitem[Wang et~al.(2023)Wang, Zhao, Sciabola, and Wang]{wang2023cmolgpt}
Wang, Y., Zhao, H., Sciabola, S., and Wang, W.
\newblock cmolgpt: A conditional generative pre-trained transformer for target-specific de novo molecular generation.
\newblock \emph{Molecules}, 28\penalty0 (11):\penalty0 4430, 2023.

\bibitem[Weiss et~al.(2016)Weiss, Bortolato, Tehan, and Mason]{weiss2016gpcr}
Weiss, D.~R., Bortolato, A., Tehan, B., and Mason, J.~S.
\newblock Gpcr-bench: a benchmarking set and practitioners’ guide for g protein-coupled receptor docking.
\newblock \emph{Journal of chemical information and modeling}, 56\penalty0 (4):\penalty0 642--651, 2016.

\bibitem[Wellawatte et~al.(2022)Wellawatte, Seshadri, and White]{wellawatte2022model}
Wellawatte, G.~P., Seshadri, A., and White, A.~D.
\newblock Model agnostic generation of counterfactual explanations for molecules.
\newblock \emph{Chemical science}, 13\penalty0 (13):\penalty0 3697--3705, 2022.

\bibitem[Wishart et~al.(2018)Wishart, Feunang, Guo, Lo, Marcu, Grant, Sajed, Johnson, Li, Sayeeda, Assempour, Iynkkaran, Liu, MacIejewski, Gale, Wilson, Chin, Cummings, Le, Pon, Knox, and Wilson]{Wishart2018}
Wishart, D.~S., Feunang, Y.~D., Guo, A.~C., Lo, E.~J., Marcu, A., Grant, J.~R., Sajed, T., Johnson, D., Li, C., Sayeeda, Z., Assempour, N., Iynkkaran, I., Liu, Y., MacIejewski, A., Gale, N., Wilson, A., Chin, L., Cummings, R., Le, D., Pon, A., Knox, C., and Wilson, M.
\newblock Drugbank 5.0: a major update to the drugbank database for 2018.
\newblock \emph{Nucleic acids research}, 46:\penalty0 D1074--D1082, 1 2018.
\newblock ISSN 1362-4962.
\newblock \doi{10.1093/NAR/GKX1037}.

\bibitem[Wognum et~al.(2024)Wognum, Ash, Aldeghi, Rodr{\'\i}guez-P{\'e}rez, Fang, Cheng, Price, Clevert, Engkvist, and Walters]{wognum2024call}
Wognum, C., Ash, J.~R., Aldeghi, M., Rodr{\'\i}guez-P{\'e}rez, R., Fang, C., Cheng, A.~C., Price, D.~J., Clevert, D.-A., Engkvist, O., and Walters, W.~P.
\newblock A call for an industry-led initiative to critically assess machine learning for real-world drug discovery.
\newblock \emph{Nature Machine Intelligence}, pp.\  1--2, 2024.

\bibitem[W{\'o}jcikowski et~al.(2017)W{\'o}jcikowski, Ballester, and Siedlecki]{wojcikowski2017performance}
W{\'o}jcikowski, M., Ballester, P.~J., and Siedlecki, P.
\newblock Performance of machine-learning scoring functions in structure-based virtual screening.
\newblock \emph{Scientific Reports}, 7\penalty0 (1):\penalty0 46710, 2017.

\bibitem[Woodhead et~al.(2024)Woodhead, Erlanson, de~Esch, Holvey, Jahnke, and Pathuri]{Woodhead2024}
Woodhead, A.~J., Erlanson, D.~A., Esch, I.~J.~de, Holvey, R.~S., Jahnke, W., and Pathuri, P.
\newblock Fragment-to-lead medicinal chemistry publications in 2022.
\newblock \emph{Journal of Medicinal Chemistry}, 67:\penalty0 2287--2304, 2 2024.
\newblock ISSN 15204804.
\newblock \doi{10.1021/ACS.JMEDCHEM.3C02070/SUPPL_FILE/JM3C02070_SI_001.PDF}.

\bibitem[Wu et~al.(2024{\natexlab{a}})Wu, Wang, Chen, Tang, Wu, and Yu]{wu2024t}
Wu, J.-N., Wang, T., Chen, Y., Tang, L.-J., Wu, H.-L., and Yu, R.-Q.
\newblock t-smiles: a fragment-based molecular representation framework for de novo ligand design.
\newblock \emph{Nature Communications}, 15\penalty0 (1):\penalty0 4993, 2024{\natexlab{a}}.

\bibitem[Wu et~al.(2024{\natexlab{b}})Wu, Xia, Deng, Liu, Zhang, Guo, Cui, Pei, Wu, Xie, et~al.]{wu2024tamgen}
Wu, K., Xia, Y., Deng, P., Liu, R., Zhang, Y., Guo, H., Cui, Y., Pei, Q., Wu, L., Xie, S., et~al.
\newblock Tamgen: drug design with target-aware molecule generation through a chemical language model.
\newblock \emph{Nature Communications}, 15\penalty0 (1):\penalty0 9360, 2024{\natexlab{b}}.

\bibitem[Xu et~al.(2021)Xu, Ran, and Chen]{xu2021novo}
Xu, M., Ran, T., and Chen, H.
\newblock De novo molecule design through the molecular generative model conditioned by 3d information of protein binding sites.
\newblock \emph{Journal of Chemical Information and Modeling}, 61\penalty0 (7):\penalty0 3240--3254, 2021.

\bibitem[Yau \& Loo(2022)Yau and Loo]{yau2022consensus}
Yau, M.~Q. and Loo, J.~S.
\newblock Consensus scoring evaluated using the gpcr-bench dataset: Reconsidering the role of mm/gbsa.
\newblock \emph{Journal of Computer-Aided Molecular Design}, 36\penalty0 (6):\penalty0 427--441, 2022.

\bibitem[Zdrazil et~al.(2024)Zdrazil, Felix, Hunter, Manners, Blackshaw, Corbett, de~Veij, Ioannidis, Lopez, Mosquera, Magarinos, Bosc, Arcila, Kizilören, Gaulton, Bento, Adasme, Monecke, Landrum, and Leach]{Zdrazil2024}
Zdrazil, B., Felix, E., Hunter, F., Manners, E.~J., Blackshaw, J., Corbett, S., Veij, M.~de, Ioannidis, H., Lopez, D.~M., Mosquera, J.~F., Magarinos, M.~P., Bosc, N., Arcila, R., Kizilören, T., Gaulton, A., Bento, A.~P., Adasme, M.~F., Monecke, P., Landrum, G.~A., and Leach, A.~R.
\newblock The chembl database in 2023: a drug discovery platform spanning multiple bioactivity data types and time periods.
\newblock \emph{Nucleic Acids Research}, 52:\penalty0 D1180--D1192, 1 2024.
\newblock ISSN 0305-1048.
\newblock \doi{10.1093/NAR/GKAD1004}.

\bibitem[Zhang et~al.(2023{\natexlab{a}})Zhang, Gao, Wang, Chen, Zhang, Chen, Li, Qi, and Wang]{zhang2023planet}
Zhang, X., Gao, H., Wang, H., Chen, Z., Zhang, Z., Chen, X., Li, Y., Qi, Y., and Wang, R.
\newblock Planet: a multi-objective graph neural network model for protein--ligand binding affinity prediction.
\newblock \emph{Journal of Chemical Information and Modeling}, 64\penalty0 (7):\penalty0 2205--2220, 2023{\natexlab{a}}.

\bibitem[Zhang et~al.(2023{\natexlab{b}})Zhang, Zhao, Xie, Bian, and Zhou]{zhang2023activity}
Zhang, Z., Zhao, B., Xie, A., Bian, Y., and Zhou, S.
\newblock Activity cliff prediction: Dataset and benchmark.
\newblock \emph{arXiv preprint arXiv:2302.07541}, 2023{\natexlab{b}}.

\bibitem[Zhou et~al.(2021)Zhou, Cao, and Skolnick]{zhou2021fragsite}
Zhou, H., Cao, H., and Skolnick, J.
\newblock Fragsite: a fragment-based approach for virtual ligand screening.
\newblock \emph{Journal of chemical information and modeling}, 61\penalty0 (4):\penalty0 2074--2089, 2021.

\bibitem[Zhou et~al.(2014)Zhou, Georgeon, Moser, Moore, Caflisch, and Hantschel]{zhou2014specificity}
Zhou, T., Georgeon, S., Moser, R., Moore, D., Caflisch, A., and Hantschel, O.
\newblock Specificity and mechanism-of-action of the jak2 tyrosine kinase inhibitors ruxolitinib and sar302503 (tg101348).
\newblock \emph{Leukemia}, 28\penalty0 (2):\penalty0 404--407, 2014.

\bibitem[Zhou et~al.(2024)Zhou, Zhang, Zhao, Yu, Shen, Zhou, Wang, Qiu, Chen, and Zhu]{Zhou2024}
Zhou, Y., Zhang, Y., Zhao, D., Yu, X., Shen, X., Zhou, Y., Wang, S., Qiu, Y., Chen, Y., and Zhu, F.
\newblock Ttd: Therapeutic target database describing target druggability information.
\newblock \emph{Nucleic acids research}, 52:\penalty0 D1465--D1477, 1 2024.
\newblock ISSN 1362-4962.
\newblock \doi{10.1093/NAR/GKAD751}.

\end{thebibliography}
\bibliographystyle{valence}  
\clearpage
\appendix
\section{Appendix}
\label{sec:appendix}

\subsection{MoAT-DB: A New Dataset of Target--Compound--MoA Relationships}
\label{appendix:dataset}

We introduce \emph{MoAT-DB}, a new dataset (\(\mathcal{D}\)) consolidating Target--Compound--MoA information from four major sources: DrugBank 5.0 \citep{Wishart2018}, Clue CMAP \citep{subramanian2017next}, TTD Database \citep{Zhou2024}, and ChEMBL34 \citep{Zdrazil2024}. These sources provide diverse annotations of molecules, targets, and mechanisms of action (MoA), differing in format and coverage. MoAT-DB standardizes these annotations to enable conditional modeling of targets, compounds, and MoAs. To ensure relevance, Target--Compound pairs in ChEMBL34 were filtered using experimental bioactivity thresholds (Table~\ref{tab:chembl_filtering}), retaining only entries with evidence of activity. Targets were identified using ChEMBL IDs and mapped to their corresponding classes via the ChEMBL protein-classification tree. In some cases, related targets were grouped into broader target families, while large families were subdivided, resulting in slight deviations from the original ChEMBL annotations. Additionally, some targets are associated with multiple protein classes, but only one was retained in MoAT-DB, which may introduce a potential limitation in classification consistency. Target variants were excluded, with ChEMBL IDs used as the primary reference.  

Since the four sources differ in how MoAs terms are reported, we manually standardized them into 22 categories. The MoA terms that are well-represented (number of occurrence is greater than 50) in the dataset include: \textit{inhibitor}, \textit{activator}, \textit{allosteric}, \textit{antagonist}, \textit{binder}, \textit{[positive/negative] modulator}, \textit{stimulator}, \textit{potentiator}, \textit{blocker}, \textit{[inverse/partial] agonist}, \textit{disruptor}, \textit{substrate}, \textit{agent}, and \textit{stabilizer}. Less common MoA terms were grouped into the \textit{other} category, and ambiguous or partially known true mechanisms of action are labeled \textit{unknown}. Cases where MoA information for molecule-target pairs was unavailable (85\% of entries), are explicitly labeled as missing information. The final dataset contains 4,010 unique targets, 39 target classes, 580,079 unique compounds, and 22 distinct MoA terms, totaling ~1.15M samples as compound-target pairs may have multiple MoAs.

\begin{figure}[tb]
\begin{center}
   \centerline{
   \includegraphics[width=\columnwidth]{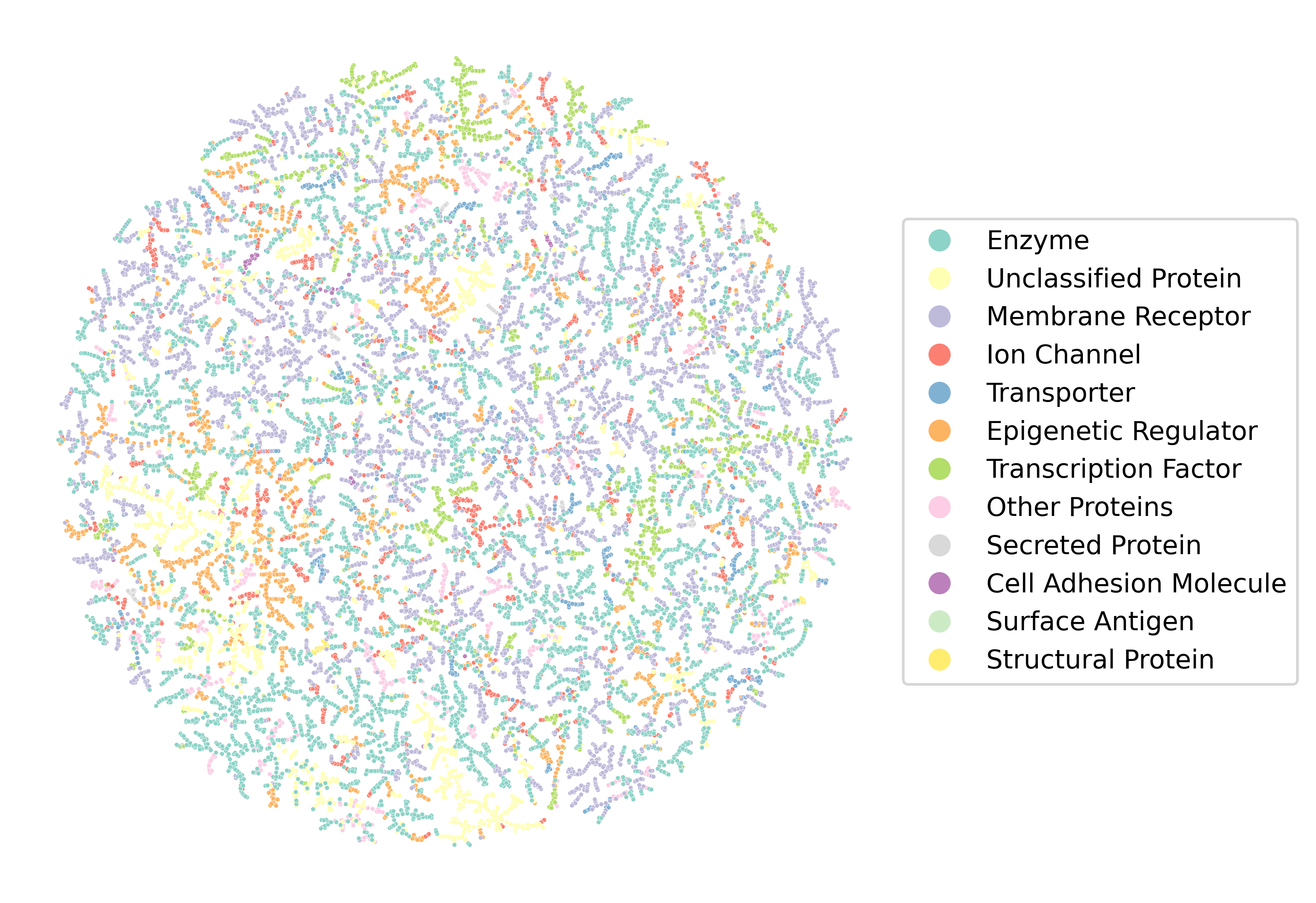}}
   \caption{TMAP visualization of chemical space coverage in MoAT-DB, colored by target family. The visualization reveals distinct clusters corresponding to different target families while showing areas of shared chemical space.}
   \label{fig:tmap}
\end{center}
\end{figure}

\begin{figure*}[!tb]
   \centering
   \includegraphics[width=\linewidth]{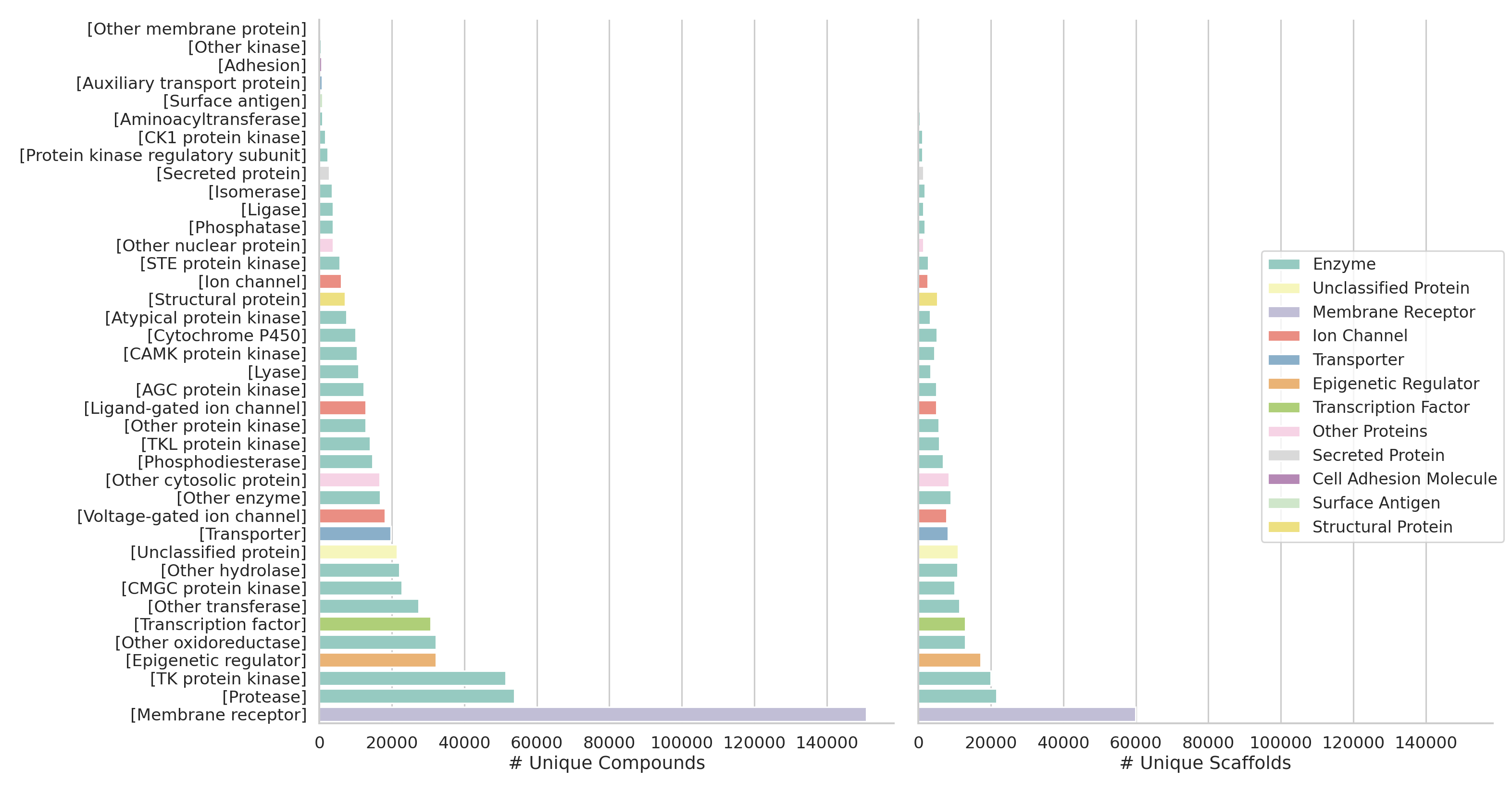}
   \caption{Distribution of unique compounds and scaffolds across target families in MoAT-DB, illustrating its chemical diversity and coverage of protein families.}
   \label{fig:dataset-cmpds-barplot}
   \vskip -0.15in
\end{figure*}

Figure~\ref{fig:tmap} provides a TMAP visualization of compounds annotated by target family and Figure~\ref{fig:dataset-cmpds-barplot} illustrates the distribution of unique molecules and scaffolds per target family, underscoring the chemical diversity within MoAT-DB.
Together, these figures offer a high-level overview of both target and molecular diversity within MoAT-DB.

\paragraph{Dataset Splits} We partition \(\mathcal{D}\) into training (\(\mathcal{D}_\text{train}\)), validation, and two out-of-distribution (OOD) test sets:
\begin{itemize}
    \item \(\mathcal{D}_\text{val}\): 10,000 molecules from the same distribution as \(\mathcal{D}_\text{train}\), stratified by Target-MoA pairs
    \item \(\mathcal{D}_\text{Mol}\): ~22,000 \emph{novel molecules} (unseen in \(\mathcal{D}_\text{train}\)) with their corresponding targets and MoAs
    \item \(\mathcal{D}_\text{MoA}\): ~5,000 samples with \emph{unseen Target--MoA combinations}, though individual targets or MoAs may appear in \(\mathcal{D}_\text{train}\)
\end{itemize}

These splits enable a more robust evaluation of generalization, both for unseen chemotypes and for previously unobserved Target--MoA assignments.

\begin{table}[H]
    \centering
   \caption{Summary of Thresholds Applied for Filtering Target--Compound--MoA Relationships}
   \label{tab:chembl_filtering}
   \begin{center}
   \begin{tabular}{ @{}lcc@{}}
   \hline
       \textbf{Endpoint} & \textbf{Unit} & \textbf{Threshold} \\ \hline
       Inhibition, Activity  & \% &  $\geq 90$ \\ 
       IC50, AC50, EC50, GI50, ED50 & nM & $\leq 1000$ \\ 
       Ki, Kd    & nM & $\leq 500$ \\ 
       \%Inhib (Mean)  & \% & $> 90$ \\ 
       Residual Activity   & \% & $\leq 10$ \\ 
       \% Control  & \% & $< 50$ \\ 
       Delta TM & C & $> 10$ \\  \hline
   \end{tabular}
   \end{center}
   \vskip -0.15in
\end{table}

\subsection{Training Processes}
\label{appendix:training_processes}

All experiments are implemented in PyTorch with Hugging Face Transformers, ensuring full reproducibility. 
The complete codebase and data preparation scripts are publicly available at: \url{https://github.com/valence-labs/SAFE-T}.

We detail below the datasets, architecture, and training procedures used to train \modelname~ models across three stages: (1) chemical pretraining, (2) biological context fine-tuning, and (3) preference optimization.

\paragraph{Training Data}
We use distinct datasets for each training phase. For pretraining, which does not involve biological conditioning, we compile a corpus of 11.4M molecules by combining in-stock, drug-like compounds from ZINC20 \citep{irwin2020zinc20} with all unique molecules from ChEMBL. For context-conditioned fine-tuning, we use MoAT-DB (see Appendix~\ref{appendix:dataset}). For preference tuning, we use pairwise tuples derived from the ACNet benchmark~\citep{zhang2023activity}.

All molecules are converted from SMILES to SAFE using the BRICS decomposition algorithm~\citep{degen2008art}. Molecules without identifiable BRICS bonds are retained as-is using canonical SMILES.

\paragraph{Biological Context Representation} For the biological context triplet \(\bigl(c_{\text{fam}}, c_{\text{tgt}}, c_{\text{moa}}\bigr)\), we encode all components using square brackets: protein families (e.g., [Kinase]), target identifiers (e.g., [CHEMBL203]), and MoA terms (e.g., [inhibitor]). We use ChEMBL IDs rather than protein sequences or structural information as they are ubiquitous in drug discovery databases and literature. These IDs offer several advantages: they not only uniquely identify individual proteins and isoforms, but also capture target complexes, gene fusions, and selectivity classes.

\paragraph{Tokenizer} We use a byte-pair encoding (BPE) tokenizer pretrained on molecular structures in SAFE format. This tokenizer initially covers structural fragments and is then expanded with 4{,}075 additional tokens corresponding to the biological context \(\bigl(c_{\text{fam}}, c_{\text{tgt}}, c_{\text{moa}}\bigr)\). The resulting vocabulary size is 5{,}070 tokens.

\paragraph{Model Variants} Following \citep{mesbahi2024safe}, we based \modelname~ models on the LLaMA architecture. In addition to the main \textbf{\modelname~} model, we further trained two smaller models (\textbf{xs-\modelname}~and \textbf{sm-\modelname}) detailed in Table~\ref{tab:model_variants}. 

\begin{table}[tb]
\caption{Summary of \modelname~ model variants}
\label{tab:model_variants}
\begin{center}
\begin{small}
\begin{tabular}{lccc}
\toprule
& \textbf{xs-\modelname~} & \textbf{sm-\modelname~} & \textbf{\modelname~} \\
\midrule
\textbf{\# Att. heads}  & 12          & 16  & 16   \\
\textbf{\# Layers}      & 6           & 8   & 16   \\
\textbf{Hidden dim}     & 384         & 512 & 512  \\
\textbf{Interm. dim}    & 512         & 1024 & 1024 \\
\textbf{\# Params}      & 10.47M      & 24.96M & 44.97M \\
\bottomrule
\end{tabular}
\end{small}
\end{center}
\vskip -0.15in
\end{table}

All training runs use up to four NVIDIA H100 GPUs with 32 CPU cores per job and a maximum runtime of 96 hours per training stage.

\subsubsection{Pre-training on Chemical Data}
The first training stage involves pre-training a generative model over SAFE strings with biological context tokens masked. This objective enables the model to capture general chemical semantics and sequence patterns, independent of biological prompts. Specifically, each input is formatted as \((c^{\text{null}}, x)\), where:
\[
c^{\text{null}} = (\text{\textless mask\textgreater}, \text{\textless mask\textgreater}, \text{\textless mask\textgreater})
\]
The model is trained with a standard causal language modeling (CLM) loss:
\[
\text{NLL}(x \mid c^{\text{null}}) = - \sum_{t=1}^{T} \log p_\theta(x_t \mid x_{<t}, c^{\text{null}})
\]
We use a cosine learning rate schedule in \(\{10^{-3}, 10^{-4}\}\), a warmup ratio of 0.1, gradient clipping at 1, and batch sizes ranging from 128 to 256, with the Adam optimizer. To improve generalization, we also randomize the ordering of SAFE fragments across batches. Input length is capped at 512 tokens. Models were trained for up to 50 epochs. 

\subsubsection{Fine-tuning with Biological Context}
\begin{figure*}[tbh]
    \centering
    \includegraphics[width=\linewidth]{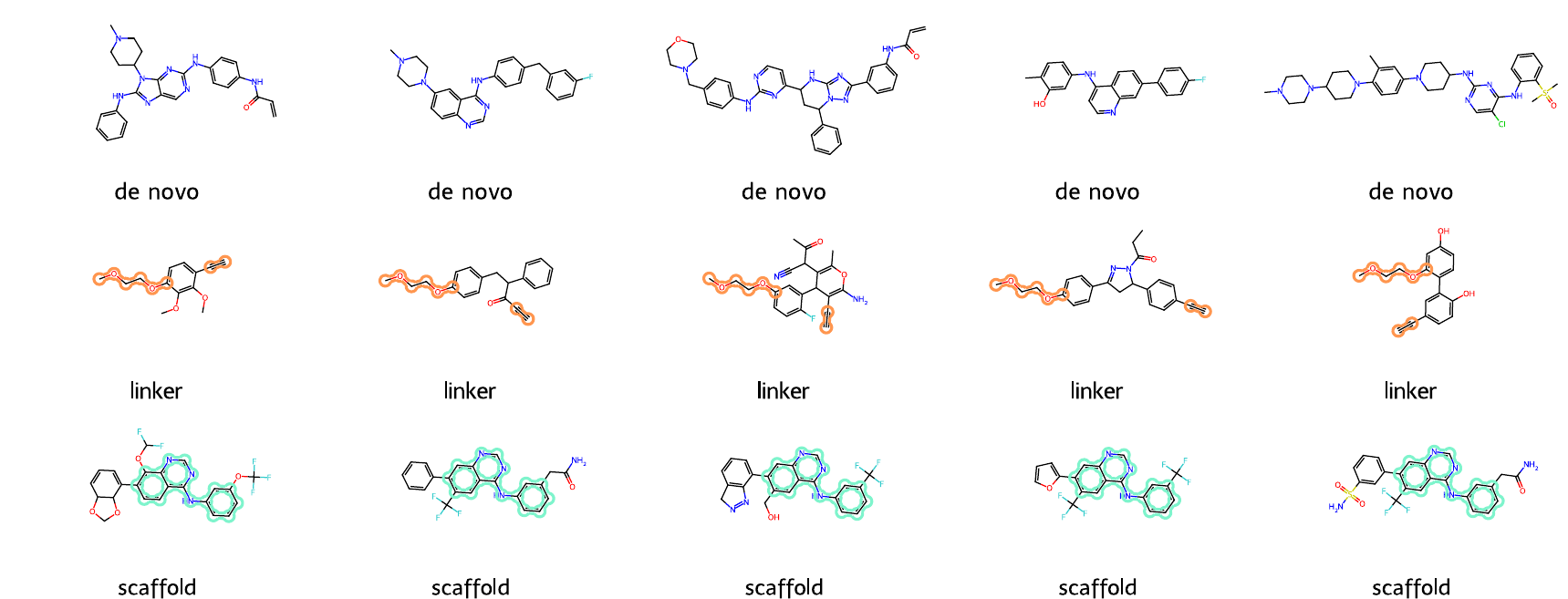}
    \small \caption{Example of sampled molecules under the \textit{de novo} and fragment-constrained design given the biological context \textit{[TK\ protein\ kinase]/[CHEMBL203]/[inhibitor]}. The shared core structures are \textit{[2*]C\#C.[14*]OCCOC} for linker design and \textit{[1*]c1cccc(Nc2ncnc3cc([2*])c([3*])cc23)c1} for scaffold decoration. The cores being decorated are highlighted in the figure.}
    \label{appendix:fig:sample-molecules}
\end{figure*}

In the second stage, the model is fine-tuned on the full MoAT-DB dataset, with biological conditions \((c_{\text{fam}}, c_{\text{tgt}}, c_{\text{moa}})\) unmasked and prepended to the molecular sequence. Random context masking (with annealed probability) is applied to improve robustness against incomplete biological conditions. All other hyperparameters remain consistent with the pretraining setup. Figure~\ref{appendix:fig:sample-molecules} highlights some example molecules generated by \modelname~ in both \textit{de novo} and structured-constrained generation. 

\subsubsection{Preference Tuning with DPO} 
The third stage calibrates the model’s likelihoods to better reflect compound ranking within each biological condition. For this, we apply Direct Preference Optimization (DPO) using pairwise preference tuples from the ACNet training dataset. Given a condition \(c^i\) and a preferred molecule \(x^{i;+}\) over \(x^{i;-}\), the DPO loss is:
\[
L_{\text{DPO}} = -\mathbb{E} \left[ \log \sigma \left( \beta \log P - \beta \log N \right) \right],
\]
\[
P = \frac{p'_\theta(x^{i;+} \mid c^i)}{p_{\text{ref}}(x^{i;+} \mid c^i)}, \quad N = \frac{p'_\theta(x^{i;-} \mid c^i)}{p_{\text{ref}}(x^{i;-} \mid c^i)}
\]
where \(p'_\theta\) is the updated model, \(p_{\text{ref}}\) is the frozen reference model from the fine-tuning stage, and \(\sigma\) is the sigmoid function. \(\beta\) controls the sharpness of the preference margin.

To ensure stability during preference tuning, we adopt slightly lower learning rates in \(\{10^{-5}, 10^{-4}\}\). This final stage aligns the model’s conditional likelihood with observed preferences and improves molecular prioritization and activity cliff detection.

\paragraph{Evaluation Metrics}

We evaluate model performance using two complementary metrics: the Area Under the Receiver Operating Characteristic Curve (ROC-AUC) for assessing ranking quality, and Top-K\% accuracy for evaluating precision in high-confidence predictions. Top-K\% accuracy measures the proportion of instances where the top-K\% predicted labels include at least one correct label. Formally, for a dataset of $N$ samples, the overall accuracy is:
\[
\text{Acc}^{\text{Top-K}}  = \frac{1}{N} \sum_{i=1}^N \text{Acc}_i^{\text{Top-K}}
\]
For a sample $i$, let $y_i \in \{0,1\}^L$ be the ground truth label vector, where $L$ is the total number of labels and $y_{i,j} = 1$ indicates relevance. Let $\hat{p}_i \in [0,1]^L$ be the predicted scores, with $\hat{p}_{i,j}$ as the score for label $j$. Define $\text{Top-K}(\hat{p}_i)$ as the indices of the top $K\%$ ranked scores. The per-sample accuracy is:
\[
\text{Acc}_i^{\text{Top-K}} = 
\begin{cases} 
1, & \text{if } y_{i,j} \in \text{Top-K}(\hat{p}_i), \\
0, & \text{otherwise}
\end{cases}
\]

\subsection{Predictive Task Formulation}
\label{appendix:predictive_task}

To investigate the utility of the \modelname~framework beyond molecular generation, we have adapted several predictive and ranking tasks relevant to early-stage drug discovery. These tasks leverage the conditional distribution \(p_\theta(x \mid c)\) over molecules \(x\), given biological contexts \(c = (c_{\text{fam}}, c_{\text{tgt}}, c_{\text{moa}})\). Through these tasks, we demonstrate that \modelname~supports biologically grounded molecular prioritization, target and MoA prediction and activity cliff detection.

\subsubsection{Biological Context Classification}
\label{appendix:generic_biocontext}

We consider the inverse problem of inferring the biological context \(c\) of a given molecule \(x\). Let \(c = (c_{\text{fam}}, c_{\text{tgt}}, c_{\text{moa}})\) be the complete context, with only a subset observed. Denote the known components as \(c_{\text{known}}\) and the unknown as \(c_{\text{unk}} \in \mathcal{C}_{\text{unk}} := \{ c_{\text{fam}}, c_{\text{tgt}}, c_{\text{moa}} \} \setminus c_{\text{known}}\), such that:
\[
c = c_{\text{known}} \cup c_{\text{unk}}.
\]
We infer the missing elements via:
\[
\hat{c}_{\text{unk}} = \argmax_{c_{\text{unk}} \in \mathcal{C}_{\text{unk}}} p(c \mid x) \propto \argmax_{c_{\text{unk}} \in \mathcal{C}_{\text{unk}}} p_\theta(x \mid c) \cdot p(c),
\]
where \(p(c)\) is a prior estimated from data or assumed uniform. To handle the computational complexity of evaluating across a large number of possible contexts, we can apply filtering based on prior knowledge (e.g., user-defined target list, suspected pathway interactions, etc.) that narrows the search space $\mathcal{C}_{\text{unk}}$.

This formulation naturally supports drug-target interaction (DTI) prediction (estimating \(c_{\text{tgt}}\) and \(c_{\text{fam}}\)) and MoA classification (\(c_{\text{moa}}\)), and allows for assessment of selectivity or polypharmacology by comparing \(\log p_\theta(x \mid c)\) across biological contexts.

\subsubsection{Activity Cliff Detection}
\label{appendix:activity_cliff}

An activity cliff is a pair of structurally similar molecules whose biological activities differ substantially. Let \((x^{(1)}, x^{(2)})\) be such a pair with minimal structural variation (e.g., differing by one or a few fragments). \modelname~ quantifies the cliff magnitude via:
\[
\Delta_{\text{cliff}}(x^{(1)}, x^{(2)}; c) = \left| \log p_\theta(x^{(1)} \mid c) - \log p_\theta(x^{(2)} \mid c) \right|.
\]
We define a cliff when \(\Delta_{\text{cliff}}\) exceeds a threshold \(\delta\), chosen empirically (e.g., via percentile-based heuristics or outlier detection). This enables interpretable cliff detection from the model’s learned conditional distribution.

\subsubsection{Target-Focused Library Design}
\label{appendix:tgt_libraries}

We use \modelname~ to design virtual screening libraries conditioned on a target and mechanism of action \(c = (c_{\text{fam}}, c_{\text{tgt}}, c_{\text{moa}})\). This can be done via:

\textbf{(i) Likelihood thresholding:}
\[
L_{\text{tgt}} = \left\{ x \mid \log p_\theta(x \mid c) > \tau \right\}
\]
where \(\tau\) is a user-defined cutoff reflecting desired affinity or biological plausibility.

\textbf{(ii) Conditional sampling:}
\[
x \sim p_\theta(x_t \mid x_{<t}, c)
\]
with optional structural scaffolds provided for fragment-constrained generation.

\begin{figure}[bt]
\begin{center}
    \includegraphics[width=\columnwidth]{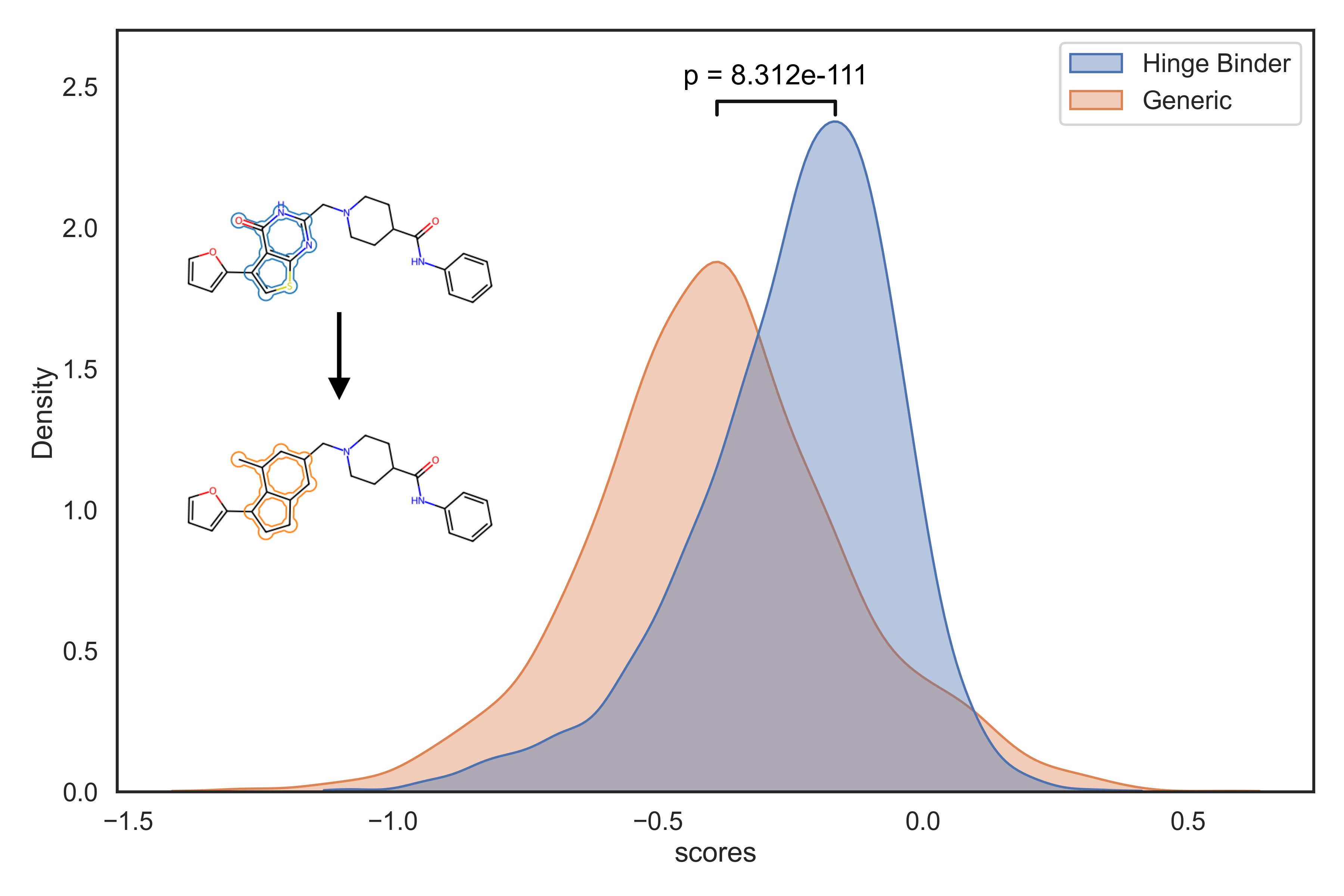}
    \caption{Likelihood distributions under a Tyrosine Protein Kinase inhibitor prompt for the Enamine Hinge Binders Library (blue) and a control library with scaffold substitution (orange). \modelname~ prioritizes kinase-relevant motifs.}
    \label{fig:hinge_binder_likelihood}
\end{center}
\end{figure}

To validate this capability, we evaluated likelihoods under a Tyrosine Protein Kinase prompt for the Enamine Hinge Binders Library and a control set with hinge motifs removed. As shown in Figure~\ref{fig:hinge_binder_likelihood}, \modelname~ assigns higher likelihoods to hinge-containing molecules, reflecting its ability to capture biologically meaningful structure–activity patterns.

\subsection{Fragment-Level Attribution for SAR Analysis}
\label{appendix:sar-analysis}

SAR analysis often hinges on understanding how individual molecular fragments contribute to biological activity. Consider a molecule \(x = \{f_1, \ldots, f_m\}\) represented as a sequence of SAFE fragments $f_i$. We can quantify the marginal attribution score for each fragment \(f_i\) under context \(c\) as:

\[
\psi_{f_i} = \log p_\theta(x \mid c) - \mathbb{E}_{f_i' \sim p_\theta(\cdot \mid x_{\setminus f_i}, c^\text{null})} \left[ \log p_\theta(x_{f_i'} \mid c) \right],
\]
where \(x_{f_i'}\) is the modified molecule with fragment \(f_i\) replaced by \(f_i'\), and \(x_{\setminus f_i}\) is the molecule with \(f_i\) removed. We sample using the null context \(c^\text{null}\) or from \modelname\textsubscript{pre} to reduce attribution bias caused by model overfitting or context amplification.

This expectation isolates the contextual importance of fragment \(f_i\) relative to plausible replacements and thus offers interpretable insights into which substructures enhance or decrease biological activity. A large \(\psi_{f_i}\) suggests that \(f_i\) contributes significantly to biological activity under \(c\). This approach provides a computationally efficient alternative to full Shapley-value analysis \citep{shapley1953value}, while maintaining a similar notion of fair attribution \citep{lundberg2017unified}.

To capture interaction effects between fragments, we can extend to pairwise contributions, albeit at a higher computational cost:
\[
\psi_{f_i f_j} = \log p_\theta(x \mid c) - \mathbb{E}_{f_i', f_j'} \left[ \log p_\theta(x_{f_i' f_j'} \mid c^\text{null}) \right] - (\psi_{f_i} + \psi_{f_j})
\]
where \(x_{f_i' f_j'} = x_{\setminus \{f_i, f_j\}} \cup \{f_i', f_j'\}\). These scores can reveal synergistic or antagonistic effects between fragment pairs, offering a higher-order view of SAR structure.

\begin{figure*}[ht]
    \centering
    \begin{minipage}{0.48\textwidth}
        \centering
        \includegraphics[width=\linewidth]{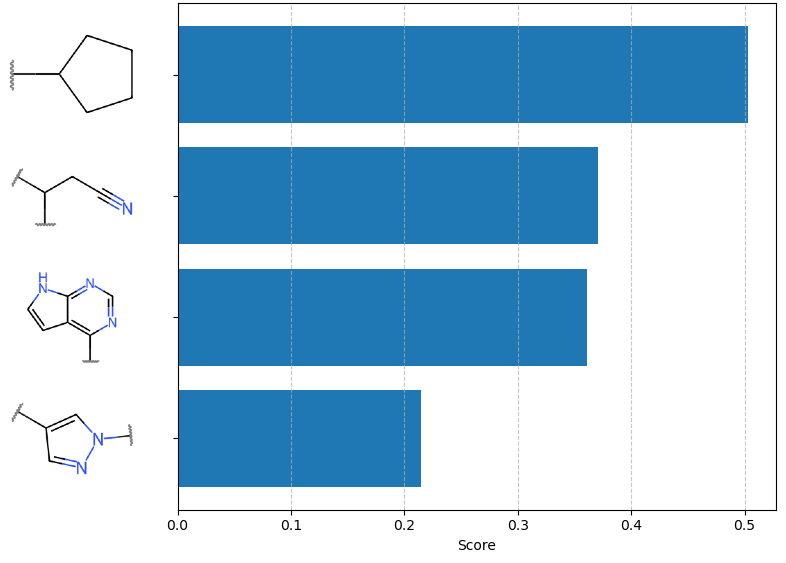}
        \label{fig:geam_jak2}
    \end{minipage}
    \hfill 
    \begin{minipage}{0.47\textwidth} 
        \centering
        \includegraphics[width=\linewidth]{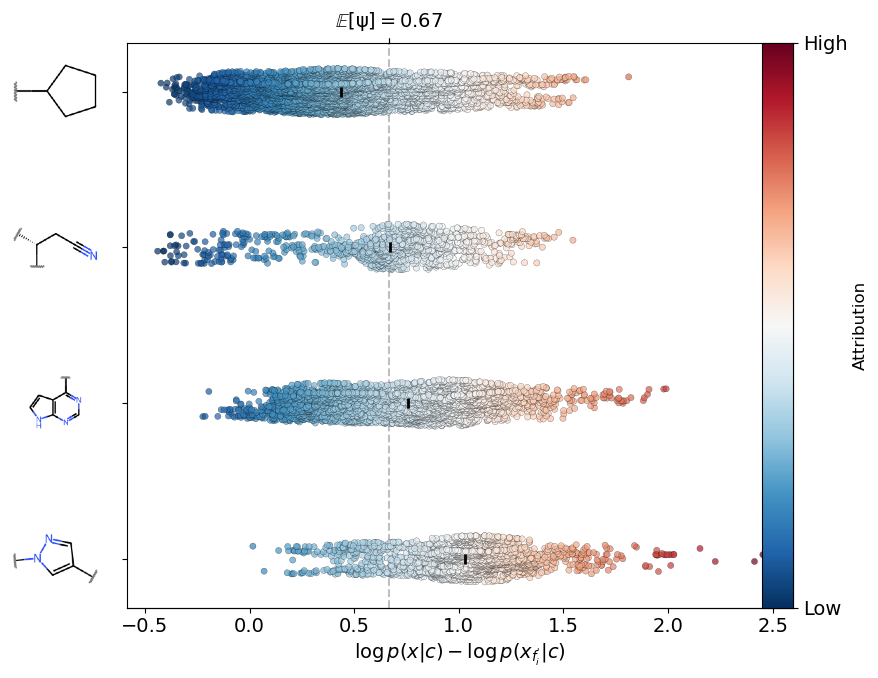}
        \label{fig:safet_jak2} 
    \end{minipage}
    
    \caption{
        \modelname~vs~GEAM fragment-level interpretability on Ruxolitinib (a JAK2 inhibitor).
        \textbf{(a)} GEAM \citep{lee2023drug} (left image) assigns the highest attribution score to the cyclopentyl group, which has been shown in the literature to be non-essential for JAK2 inhibitory activity. 
        \textbf{(b)} \modelname~(right image) correctly identifies more relevant fragments (e.g the central pyrrolopyrimidine core) that aligns with known structure-activity relationship studies of Ruxolitinib \citep{zhou2014specificity,davis2021structural}. 
        \label{fig:jak2_comparison_overall}
    }
\end{figure*}

\autoref{fig:jak2_comparison_overall} complements the main text attribution analysis by comparing \modelname~to GEAM~\citep{lee2023drug} on Ruxolitinib, another JAK inhibitor. While GEAM emphasizes less critical groups, \modelname~correctly assigns high attribution to the known hinge-binding core, demonstrating its understanding of SAR.

\paragraph{Counterfactual SAR Exploration.} While the approach described above helps understand fragment importance, it does not directly suggest structural modifications that could improve bioactivity \citep{wellawatte2022model}. Counterfactual analysis addresses this by identifying minimal changes that significantly alter predictions \citep{wachter2017counterfactual}. We generate counterfactual molecules \( x' \) through targeted fragment substitution. Given \(x = \{f_1, \ldots, f_m\}\), we replace fragment \(f_i\) with \(f_i'\) drawn from a distribution conditioned on any context \(c'\), and solve:
\[
x' = x_{\setminus f_i} \cup \{f_i'\}, \quad f_i' \sim p_\theta(\cdot \mid x_{\setminus f_i}, c')
\]
\[
\text{minimize } d(x, x') \quad \text{s.t.} \quad \left| \log p_\theta(x' \mid c) - \log p_\theta(x \mid c) \right| \ge \delta
\]
where \(d(\cdot, \cdot)\) measures molecular similarity (e.g., Tanimoto), and \(\delta\) specifies a minimum activity shift. This strategy resembles matched molecular pair analysis~\citep{hussain2010computationally} and enables hypothesis generation for lead optimization.

\begin{figure}[tb]
\begin{center}
    \includegraphics[width=0.85\linewidth]{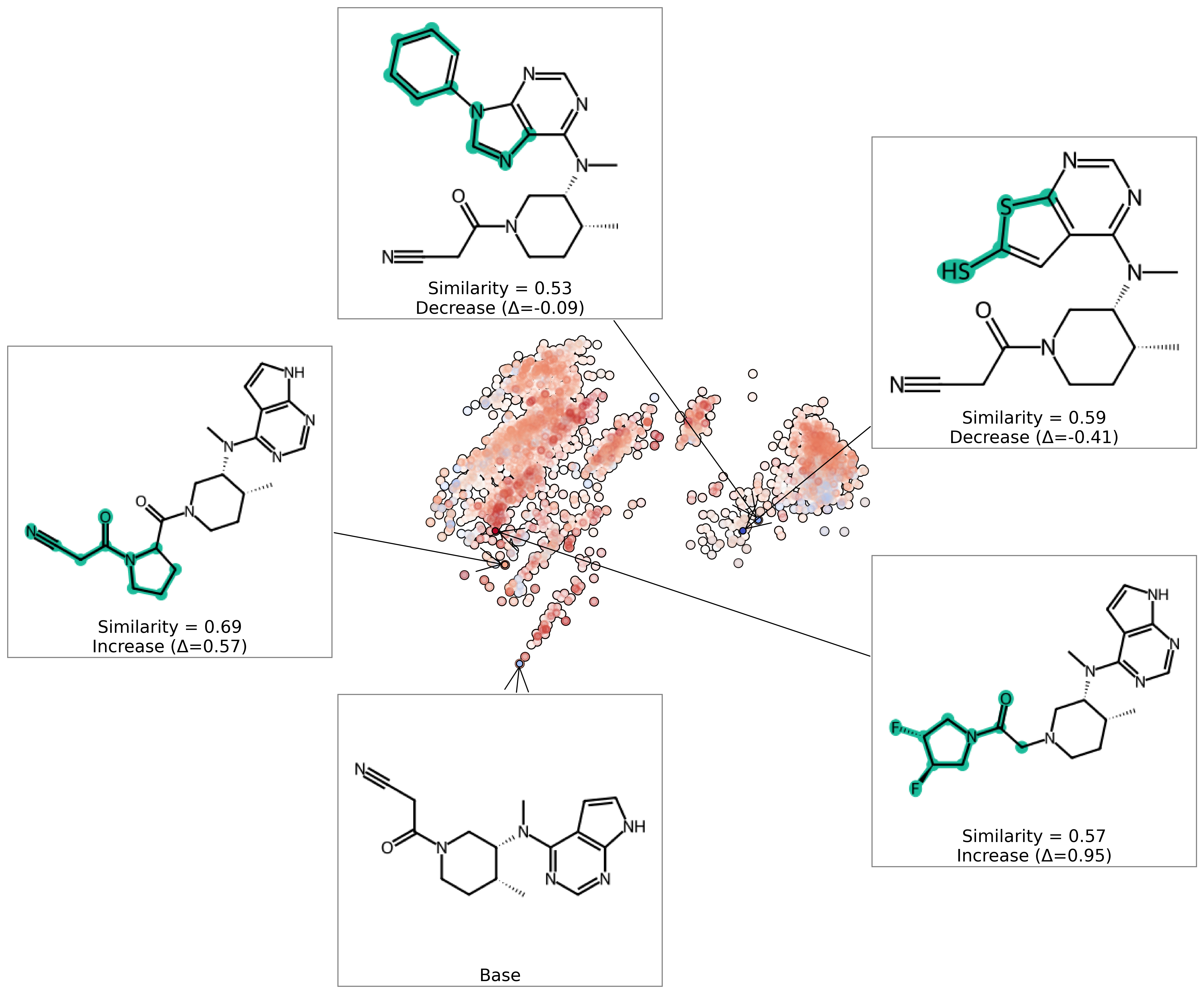}
    \caption{Counterfactual space around Tofacitinib (JAK1 inhibitor). $\Delta$ represents the difference in likelihood under true context \(c^\text{true}\), between the counterfactual $x'$ molecule and the original $x$ molecule. Positive $\Delta$  (orange) indicates changes that improve bioactivity, while negative $\Delta$ (blue) indicates the opposite.}
    \label{fig:counterfactual_kinase}
\end{center}
\end{figure}

In Figure~\ref{fig:counterfactual_kinase}, we illustrate a counterfactual exploration around Tofacitinib, a known Janus kinase (JAK) inhibitor. Fragment replacements are sampled either from the true distribution \(p_\theta(\cdot \mid x_{\setminus f_i}, c^\text{true})\) or a random or unconditioned one, e.g. \(p_\theta(\cdot \mid x_{\setminus f_i}, c^\text{null})\). Conditioning on \(c^\text{true}\) yields fragments more likely to preserve critical target interactions, whereas \(c^\text{null}\) induce broader structural variations. Mapping these substitutions into chemical space, annotated by the ground-truth conditional likelihood, shows how \modelname~ captures essential structure-activity relationships and reveals how variation in molecular substructures influences biological fitness.

\subsection{Normalization Techniques for Likelihood Estimation}
\label{appendix:lkl_normalization}

To accurately assess molecular likelihoods under specified biological conditions, normalization techniques might be essential to mitigate biases arising from dataset distributions, prior conditions, or variations in model outputs. Below, we describe three normalization approaches: (1) normalization by a generated background population, (2) normalization by null condition, and (3) Bayesian adjustment using condition priors. These techniques were evaluated across the various test sets to assess their performance using Top-1\% Accuracy and ROC-AUC metrics.

\paragraph{Normalization by Background Population ($l_{pop}$)}
This approach evaluates the likelihood of a molecule \( x \) relative to a reference distribution of molecules generated under the same condition \( c^i \). The normalized score is computed as:
\[
l_{pop}(x, c^i) = \log p(x | c^i) - \mathbb{E}[\log p(\tilde{x} | c^i)]
\]
where \( \tilde{x} \) represents a background population generated under \( c^i \), and \( \mathbb{E}[\log p(\tilde{x} | c^i)] \) is the expected likelihood of this population. This normalization effectively calibrates molecular likelihoods within the same condition, making it well-suited for molecular prioritization tasks such as virtual screening, where ranking molecules by their relative fit within a biological context is critical. However, its effectiveness depends on the diversity and representativeness of the background population.

\paragraph{Normalization by Null Condition ($l_{null}$)}:
This approach contrasts the likelihood of \( x \) under \( c^i \) with its likelihood in the absence of any biological conditioning, \( c^\text{null} = (\text{\textless mask\textgreater}, \text{\textless mask\textgreater}, \text{\textless mask\textgreater}) \), yielding the normalized score:
\[
l_{null}(x, c^i) = \log p(x | c^i) - \log p(x | c^\text{null})
\]
By isolating the informational gain provided by \( c^i \), this method is well-suited for context classification tasks. This makes \( l_{null} \) particularly valuable for tasks such as target and MoA classification, where the goal is to assign molecules to their most probable biological contexts rather than rank them within a fixed condition.

\paragraph{Bayesian Normalization with Condition Prior ($l_{prior}$)}: This approach incorporates the prior probability of each condition \( p(c_i) \) computed from the training set to adjust the likelihood. The score becomes:
\[
l_{prior}(x, c^i) = \log p(x | c^i) + \log p(c^i)
\]
This method accounts for dataset biases but may penalize underrepresented conditions in scenarios requiring extrapolation.

\paragraph{Combining Bayesian Prior and Other Normalizations}: $l_{null}$ and $l_{pop}$ can be combined with the Bayesian prior, resulting in the following normalized scores:
\begin{align*}
l_{null+prior} &= \log p(x | c^i) - \log p(x | c^\text{null}) + \log p(c^i) \\
l_{pop+prior} &= \log p(x | c^i) - \mathbb{E}[\log p(\tilde{x} | c^i)] + \log p(c^i)
\end{align*}

\paragraph{Comparison of Normalization Techniques}: we evaluated our various normalization strategies, including the vanilla likelihood ($l$),  across two key drug discovery tasks: molecular prioritization, which aims to rank compounds by their predicted activity against a target, and drug-target interaction (DTI) classification, which predicts whether a compound will interact with a specific protein target. Our evaluation revealed distinct optimal strategies for each task.

For molecular prioritization, population-based normalization ($l_{pop}$) demonstrates superior performance in OOD scenarios. This approach normalizes each molecule's likelihood score against the expected likelihood of a reference population sampled under the same biological context, effectively capturing relative rankings by quantifying deviations from context-conditional distributions. In contrast, null conditioning ($l_{null}$), while excelling on IID validation, shows degraded performance on OOD tests, suggesting limitations in generalizing beyond training distributions. The pattern reverses for DTI classification, where $l_{pop}$ consistently underperforms while vanilla likelihood ($l$) and null conditioning ($l_{null}$) achieve the best results. This suggests that DTI classification benefits more from preserving the original likelihood structure or contrasting against an unconditional baseline rather than population-level statistics.
Interestingly, combining strategies with Bayesian prior (e.g., $l_{pop+prior}$ or $l_{null+prior}$) generally does not yield consistent improvements across all tasks. We hypothesize that while prior adjustment ($l_{prior}$) helps mitigate dataset biases, it introduces limitations in generalizing to rare conditions, leading to degraded performance in extrapolation tasks. These findings suggest a clear task dependence in normalization effectiveness: molecular prioritization benefits from population-level statistics that emphasize relative rankings, while DTI classification performs better with simpler approaches that preserve likelihood structure. We therefore recommend task-specific normalization selection: $l_{pop}$ for molecular prioritization and either $l$ or $l_{null}$ for DTI classification.

\begin{table}[bt]
\caption{Effect of likelihood normalization strategies on molecule prioritization and DTI classification performance (Top-1\% Accuracy). Bold values indicate best performance for each task, while second best performance is underlined.}
\label{tab:normalization_comparison}
\begin{center}
\begin{threeparttable}
\begin{tabular}{lcccccc}
\toprule
\multirow{2}{*}{\makecell[l]{Normalization\\Strategy}} & \multicolumn{3}{c}{Mol Prioritization} & \multicolumn{3}{c}{DTI Classification} \\
\cmidrule(lr){2-4} \cmidrule(lr){5-7}
& $\mathcal{D}_\text{val}$ & $\mathcal{D}_\text{Mol}$ & $\mathcal{D}_\text{MoA}$ & $\mathcal{D}_\text{val}$ & $\mathcal{D}_\text{Mol}$ & $\mathcal{D}_\text{MoA}$ \\
\midrule
$l$              & 0.533          & \underline{0.787} & 0.633          & 0.765          & \textbf{0.892} & 0.621          \\
$l_\text{prior}$      & 0.534          & 0.766          & \underline{0.658} & \textbf{0.775} & 0.889 & 0.619          \\
\midrule
$l_\text{pop}$        & 0.554          & \textbf{0.792} & \textbf{0.667} & 0.691          & 0.813          & 0.489          \\
$l_\text{pop+prior}$  & 0.524          & 0.772          & 0.635          & 0.697          & 0.816          & 0.481          \\
\midrule
$l_\text{null}$       & \textbf{0.785} & 0.742          & 0.621          & \underline{0.771} & 0.887          & \underline{0.626}          \\
$l_\text{null+prior}$ & \underline{0.779} & 0.729          & 0.604          & 0.768          & \underline{0.891}          & \textbf{0.628} \\
\bottomrule
\end{tabular}
\end{threeparttable}
\end{center}
\vskip -0.1in
\end{table}

\subsection{Details on Virtual Screening Benchmark}
\label{appendix:mol_prioritization}

\paragraph{Datasets:} We evaluate \modelname~ on molecule prioritization using four established virtual screening benchmarks: GPCR-Bench~\citep{weiss2016gpcr}, LIT-PCBA~\citep{tran2020lit} and DUD-E~\citep{mysinger2012dude}. Each benchmark presents distinct challenges and validation scenarios. \textbf{DUD-E} provides computationally generated decoys designed to challenge molecular docking programs while maintaining similar physical property distributions to known actives.
\textbf{GPCR-Bench} specializes in G protein-coupled receptors, offering a dataset of 24 targets that integrates ChEMBL bioactivity data with carefully selected decoys for evaluating deep learning models.
\textbf{LIT-PCBA} is a highly curated benchmark from PubChem bioassays, incorporating stringent filtering criteria (0.5 $<$ Hill slope $<$ 2.0) to ensure high-quality dose-response data. It poses challenges due to class imbalance, low-potency actives, and the inclusion of diverse assay types, making it a valuable benchmark for both ligand-based and structure-based models. 
While there is potential data overlap with our training set (see Table~\ref{tab:mol_prior_dataset_stats}), we report performance on the full datasets following standard practices in virtual screening benchmarking. 

\begin{table}[tb]
    \centering
    \caption{Dataset statistics and performance metrics for molecular prioritization benchmarks. Dataset overlap percentages with MoAT-DB training set indicate potential evaluation biases.}
    \label{tab:mol_prior_dataset_stats}

    \begin{tabular}{l c c c}
        \toprule
        \textbf{Dataset} & \textbf{LIT-PCBA} & \textbf{DUD-E} & \textbf{GPCR-Bench} \\
        \midrule
     \multicolumn{4}{c}{\textbf{Dataset Statistics}} \\
        \midrule
        \# Targets                      & 15         & 102      & 24       \\
        \# Actives                      & 7,844   & 22,886   & 4,264    \\
        \# Decoys              & ~407K & ~1.178M & ~250K \\
        \midrule
        \multicolumn{4}{c}{\textbf{Overlap with MoAT-DB}} \\
        \midrule
        Cmpds (\%)       & 6.43  & 1.37  & 1.43  \\
        Active Cmpds (\%)       & 14.70   & 74.81  & 94.62  \\
        Cmpds-Target (\%) & 4.91   & 69.54 & 85.57 \\

        \bottomrule
    \end{tabular}
    \vskip -0.15in
\end{table}

\paragraph{Evaluation Metrics:} Following standard protocols in virtual screening~\citep{truchon2007evaluating}, we report the Area Under the ROC Curve (\textbf{ROC-AUC}) and Enrichment Factor at 1\% (\textbf{EF@1\%}). The Enrichment Factor measures early recognition performance, quantifying how well a model ranks active compounds in the top fraction of screened molecules:
\begin{equation*}
    EF_{\alpha} = \frac{NTB_{\alpha}}{NTB_t \times \alpha}
\end{equation*}
where \( NTB_{\alpha} \) is the number of actives in the top \( \alpha \)\% of the ranked list, \( NTB_t \) is the total number of actives, and \( \alpha = 1\%\).

\paragraph{Inference Speed Benchmark.}
\modelname~achieves high-throughput inference suitable for large-scale virtual screening. Table~\ref{tab:screening_throughput} reports the inference throughput for the largest model variant on both CPU (16 AMD EPYC 7742 cores) and a single NVIDIA A100 GPU. On GPU, throughput reaches nearly 800 molecules per second, enabling screening of the entire ChEMBL library (over 2 million compounds) in under one hour per biological context. In contrast, GNINA screens approximately 240 molecules per hour on GPU~\citep{mcnutt2021gnina}, requiring more than a year to process ChEMBL. DrugCLIP and Planet, both deep learning-based scoring models, also require several hours to days per target~\citep{gao2024drugclip, zhang2023planet}, making them significantly slower for real-time screening workflows. These results underscore \modelname's efficiency as a practical tool for prioritization in early drug discovery.

\begin{table}[htb]
\centering
\caption{Inference throughput of \modelname~(largest model variant) on CPU and GPU. Values are averaged across 10 runs on ChEMBL-sized samples.}
\label{tab:screening_throughput}
\resizebox{\linewidth}{!}{
\begin{tabular}{lccccc}
\toprule
\textbf{Batch Size} & \textbf{Tokens/s (CPU)} & \textbf{Tokens/s (GPU)} & \textbf{Molecules/s (CPU)} & \textbf{Molecules/s (GPU)} & \textbf{GPU Speedup} \\
\midrule
128 & 1,965 & 119,699 & 13.01 & 792.71 & 60.9$\times$ \\
256 & 2,042 & 138,361 & 11.80 & 799.77 & 67.8$\times$ \\
512 & 2,216 & 136,368 & 11.14 & 685.27 & 61.5$\times$ \\
\bottomrule
\end{tabular}
}
\vspace{0.5cm}
\end{table}

\paragraph{Performance on LIT-PCBA, DUD-E and GPCR-Bench}
Table~\ref{appendix:tab:pcba-gpcr} presents the performance comparison of \modelname~ against multiple baselines on key virtual screening benchmarks. \modelname~ achieves strong results across LIT-PCBA, DUD-E, and GPCR-Bench, surpassing most baseline methods. However, we note that both DUD-E and GPCR-Bench contain significant data overlap with \modelname~’s training set, which may contribute to its high performance on these benchmarks. To provide a more rigorous comparison, we include RFScore-VS \citep{wojcikowski2017performance}, a model trained specifically on DUD-E. While \modelname~ still achieves the highest scores, this overlap should be considered when interpreting the results, as it may inflate performance relative to LIT-PCBA's results.

\begin{table}[tb]
    \caption{Mean performance on Virtual Screening Benchmarks}
    \label{appendix:tab:pcba-gpcr}
    \centering
    \setlength{\tabcolsep}{4pt} 
    \begin{threeparttable}  
    \begin{tabular}{llcc}
    \toprule
    \multirow{2}{*}{\textbf{Dataset}} & \multirow{2}{*}{\textbf{Method}} & 
    \multirow{2}{*}{\textbf{ROC-AUC}} &
    \multirow{2}{*}{\textbf{EF@1\%}} \\\\
    \midrule
    \multirow{10}{*}{LIT-PCBA} 

    & 2D ECFP4 search\tnote{a} & - & 2.49 \\ 
    & FragSite (\footnotesize{\citeauthor{zhou2021fragsite}})\tnote{a} & - & 4.78 \\ 
    & Surflex (\footnotesize{\citeauthor{spitzer2012surflex}})\tnote{b}  & 0.51 & 2.50  \\
    & Glide-SP (\footnotesize{\citeauthor{halgren2004glide}})\tnote{b} & 0.53 & 3.41  \\ 
    & Gnina (\footnotesize{\citeauthor{mcnutt2021gnina}})\tnote{b} & \textbf{0.61} & 4.63  \\ 
    & $\Delta$vinaRF\textsubscript{20}(\footnotesize{\citeauthor{wang2017improving}})\tnote{a}  & - & 5.38  \\
    & DeepDTA (\footnotesize{\citeauthor{ozturk2018deepdta}})\tnote{b} & 0.56 & 1.47  \\ 
    & BigBind (\footnotesize{\citeauthor{brocidiacono2023bigbind}})\tnote{b} & \textbf{0.61} & 3.82  \\ 
    & Planet (\footnotesize{\citeauthor{zhang2023planet}})\tnote{b} & 0.57 & 3.87  \\ 
    & DrugCLIP (\footnotesize{\citeauthor{gao2024drugclip}})\tnote{b} & 0.57 & 5.51  \\ 
    & \textbf{\modelname~ (\footnotesize{Ours})} & 0.57 & \textbf{6.63}  \\ 


    \midrule
    \multirow{2}{*}{DUD-E}  
    & LigMate  (\footnotesize{\citeauthor{Grimm2020}}) & 0.81 & 36.14  \\
    & Vina (\footnotesize{\citeauthor{trott2010autodockvina}})\tnote{d} & 0.73 & 9.93  \\
    & Vinardo (\footnotesize{\citeauthor{sunseri2021virtual}})\tnote{d} & 0.75 & 12.4  \\
    & Gnina (\footnotesize{\citeauthor{sunseri2021virtual}})\tnote{d} & 0.77 & 20.9  \\
    & RFScore-VS (\footnotesize{\citeauthor{wojcikowski2017performance}})\tnote{d} & 0.94 & 49.3  \\
    & Dock (\footnotesize{\citeauthor{ewing2001dock}}) & - & 32.05  \\ 
    & \textbf{\modelname~ (\footnotesize{Ours})} & \textbf{0.97} & \textbf{58.66}  \\ 
 
    \midrule
    \multirow{7}{*}{GPCR-Bench} 
    & Glide-SP (\footnotesize{\citeauthor{halgren2004glide}})\tnote{c} & - & 12.58  \\ 
    & AutoDockGPU (\footnotesize{\citeauthor{santos2021autodock}})\tnote{c} & - & 5.38  \\ 
    & Vina (\footnotesize{\citeauthor{trott2010autodockvina}})\tnote{c} & - & 5.32  \\
    & $\Delta$vinaXGB (\footnotesize{\citeauthor{wang2017improving}})\tnote{c}  & - & 0.7  \\
    & DSX (\footnotesize{\citeauthor{neudert2011dsx}})\tnote{c} & - & 4.18  \\ 
    & Dock (\footnotesize{\citeauthor{ewing2001dock}})\tnote{c} & - & 2.68  \\ 
    & MM/GBSA\tnote{c} & - & 8.70  \\ 
    & \textbf{\modelname~ (\footnotesize{Ours})} & \textbf{0.92} & \textbf{57.53}  \\ 
    \bottomrule
     
    \end{tabular}
    \begin{tablenotes}
        \item [a] as reported in \cite{zhou2021fragsite}
        \item [b] as reported in \cite{gao2024drugclip}
        \item [c] as reported in \cite{yau2022consensus}
        \item [d] as reported in \cite{sunseri2021virtual}
    \end{tablenotes}
    \end{threeparttable}
    \vskip -0.15in
\end{table}

\subsection{Additional Results and Ablation Studies}
\label{appendix:ablation}

Here we perform an ablation to dissect the influence of model size and training stage on four key tasks including molecular generation, biological context prediction, molecular prioritization and activity prediction tasks. By systematically varying model size and training stage, we aim to gain a deeper understanding of their individual and combined effects on performance. 

\begin{figure}[tb]
    \centering
    \includegraphics[width=\columnwidth]{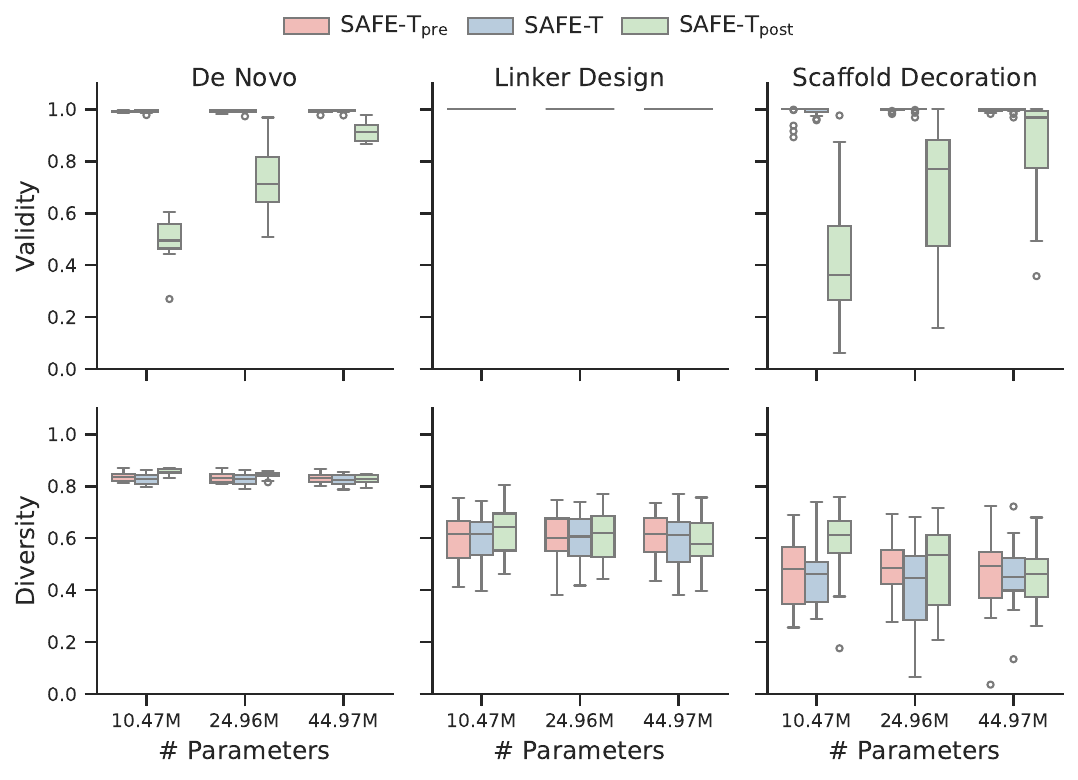}
    \small \caption{Distribution of generative design performance as a function of \modelname~ model size and training stage.}
    \label{fig:generative-metrics}
\end{figure}

\paragraph{Conditional Molecule Generation} We evaluated \modelname~ models on \textit{de novo} generation, linker design, and scaffold decoration. As shown in Figure~\ref{fig:generative-metrics}, validity and diversity remain high across tasks and model sizes, though diversity decreases slightly in fragment-constrained tasks, likely due to structural restrictions.

A notable trend is the loss of validity in preference-tuned models (\modelname$_{post}$), with smaller models struggling to maintain high validity. However, this effect diminishes as model size increases, with the largest \modelname$_{post}$ models recovering validity across all tasks. This pattern indicates that preference tuning introduces constraints that initially hinder molecular feasibility but can be compensated for by increasing model capacity. Overall, this highlights a trade-off between fine-tuning for molecular properties and maintaining validity, with scaling helping to restore performance.

\paragraph{Goal-Directed Molecular Optimization}
To complement the results in Section~\ref{sec:goal-directed-opt}, we report additional properties of molecules generated by \modelname~for the PMO benchmark tasks. In addition to achieving state-of-the-art oracle scores, the model produces chemically diverse and synthetically accessible molecules across all targets (Table~\ref{tab:pmo_properties}). This indicates that \modelname~does not simply optimize for activity but also maintains desirable structural and practical lead-like attributes.

\begin{table}[h]
\centering
\caption{Additional properties of molecules generated by \modelname~ for the PMO benchmark tasks. Diversity is measured as the pairwise Tanimoto similarity complement. SA score refers to the Synthetic Accessibility score (lower is better).}
\label{tab:pmo_properties}
\begin{tabular}{lccc}
\toprule
\textbf{Metric} & \textbf{DRD2} & \textbf{JNK3} & \textbf{GSK-3$\beta$} \\
\midrule
Diversity & 0.847 $\pm$ 0.001 & 0.857 $\pm$ 0.001 & 0.872 $\pm$ 0.001 \\
SA score & 2.741 $\pm$ 0.006 & 2.765 $\pm$ 0.014 & 2.699 $\pm$ 0.012 \\
\bottomrule
\end{tabular}
\end{table}

\begin{figure}[h]
    \centering
    \includegraphics[width=0.6\linewidth]{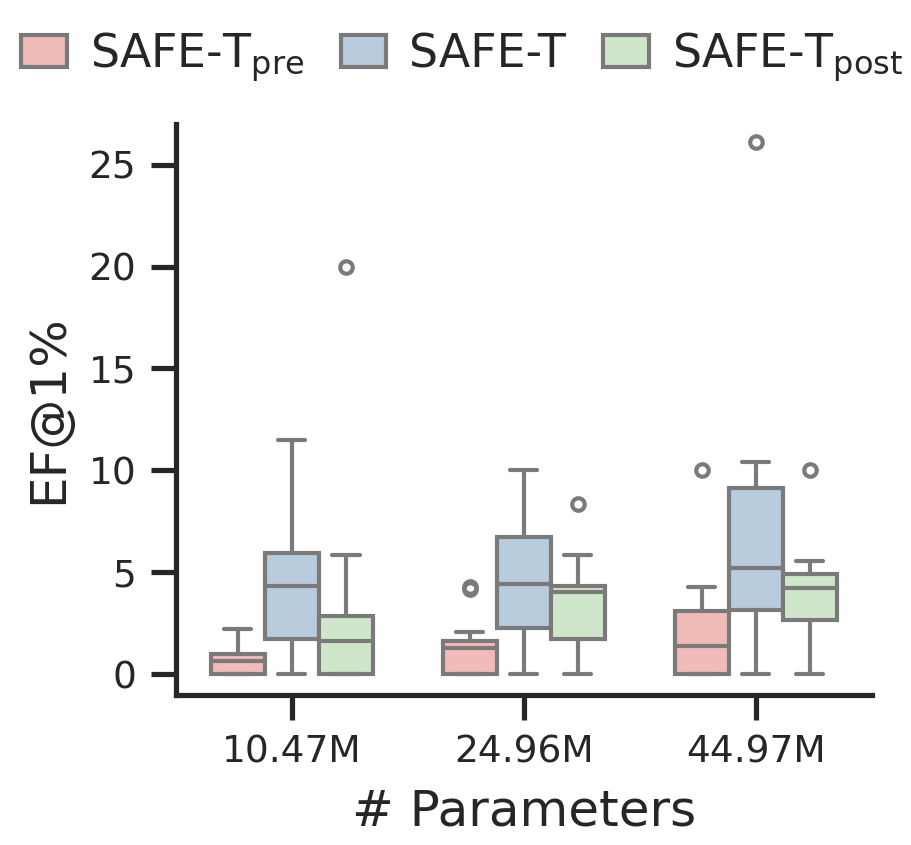}
    \small \caption{Performance distribution of \modelname~ models in molecule prioritization on the LIT-PCBA benchmark (15 targets) as a function of model size and training stage.}
    \label{appendix:mol_prio_ablation_fig}
\end{figure}

\paragraph{Molecular Prioritization}
We examine the effect of model size and training stage on molecular prioritization using the LIT-PCBA benchmark (Figure~\ref{appendix:mol_prio_ablation_fig}). As expected, larger models consistently outperform smaller ones across all training stages. Fine-tuned \modelname~ models also show superior ranking performance compared to pre-trained and preference-tuned variants.  

\begin{figure}[h]
    \centering
    \includegraphics[width=0.6\linewidth]{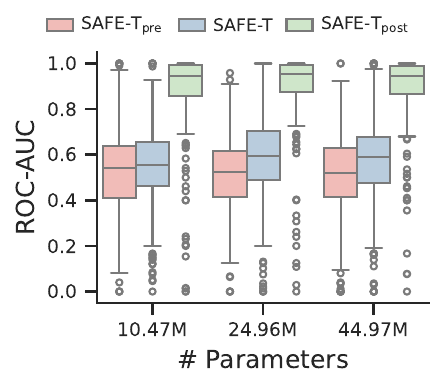}
    \small \caption{Performance of \modelname~ models on activity cliff prediction across 185 Target-MoA combinations in the ACNet dataset, evaluated at various training stages.}
    \label{appendix:ac_ablation_fig}
\end{figure}

\paragraph{Activity Cliff Prediction}  
We evaluate the impact of preference fine-tuning on activity cliff prediction. Figure~\ref{appendix:ac_ablation_fig} shows the performance distribution across 185 Target-MoA combinations in the ACNet dataset. Preference tuning drives most of the improvement, while model size has a minor effect. Although larger models generally perform better in our ablation studies, their influence is less pronounced in this task, emphasizing the importance of explicit supervision for activity cliff prediction.

\begin{table*}[!h]
\caption{Comparison of target prediction performance with and without MoA information across different datasets and model variants. For the ``With MoA'' setting, targets are predicted by marginalizing over target-MoA pairs. The baseline MolGPS+MLP is compared against \modelname~ models of increasing size. Best results for each dataset and metric are shown in \textbf{bold}.}
\label{appendix:tab:target_prediction}
\centering
\begin{small}
\begin{tabular}{lcccccc}
\toprule
\multirow{2}{*}{\textbf{Dataset}} & \multirow{2}{*}{\textbf{Model}} & \multirow{2}{*}{\textbf{\# Params}} & \multicolumn{2}{c}{\textbf{Without MoA}} & \multicolumn{2}{c}{\textbf{With MoA}} \\
\cmidrule(lr){4-5} \cmidrule(lr){6-7}
& & & \textbf{ACC@1\%} & \textbf{ROC-AUC} & \textbf{ACC@1\%} & \textbf{ROC-AUC} \\
\midrule
\multirow{4}{*}{$\mathcal{D}_\text{val}$} 
& MolGPS+MLP & - & 0.644 & 0.961 & - & - \\
& \modelname~ & 10.47M & 0.719 & 0.960 & 0.743 & 0.960 \\
& \modelname~ & 24.96M & 0.733 & 0.970 & 0.764 & 0.970 \\
& \modelname~ & 44.97M & 0.765 & 0.975 & \textbf{0.771} & \textbf{0.980} \\
\midrule
\multirow{4}{*}{$\mathcal{D}_\text{Mol}$}
& MolGPS+MLP & - & 0.792 & \textbf{0.960} & - & - \\
& \modelname~ & 10.47M & 0.862 & 0.890 & 0.876 & 0.910 \\
& \modelname~ & 24.96M & 0.871 & 0.910 & \textbf{0.887} & 0.920 \\
& \modelname~ & 44.97M & 0.870 & 0.920 & 0.885 & 0.920 \\
\midrule
\multirow{4}{*}{$\mathcal{D}_\text{MoA}$}
& MolGPS+MLP & - & 0.413 & 0.917 & - & - \\
& \modelname~ & 10.47M & 0.499 & 0.920 & 0.567 & 0.910 \\
& \modelname~ & 24.96M & 0.496 & 0.920 & 0.591 & 0.920 \\
& \modelname~ & 44.97M & 0.560 & 0.923 & \textbf{0.626} & \textbf{0.930} \\
\midrule
\multirow{4}{*}{DAVIS}
& MolGPS+MLP & - & \textbf{0.574} & 0.607 & - & - \\
& \modelname~ & 10.47M & 0.441 & 0.497 & 0.456 & \textbf{0.620} \\
& \modelname~ & 24.96M & 0.397 & 0.469 & 0.456 & 0.603 \\
& \modelname~ & 44.97M & 0.426 & 0.464 & 0.368 & 0.610 \\
\midrule
\multirow{4}{*}{KIBA}
& MolGPS+MLP & - & 0.456 & \textbf{0.813} & - & - \\
& \modelname~ & 10.47M & 0.560 & 0.739 & 0.410 & 0.702 \\
& \modelname~ & 24.96M & 0.584 & 0.790 & 0.424 & 0.719 \\
& \modelname~ & 44.97M & \textbf{0.713} & 0.791 & 0.486 & 0.727 \\
\bottomrule
\end{tabular}
\end{small}
\end{table*}

\paragraph{Drug-Target Interaction Prediction}
\label{appendix:dti-prediction}
We investigated how model size and the incorporation of mechanism of action (MoA) information influence the performance of drug-target interaction prediction. We compared two prediction approaches: (1) direct target prediction without MoA (thus masked): \emph{Without MoA}; and (2) prediction incorporating the MoA information: \emph{With MoA}. As shown in Table~\ref{appendix:tab:target_prediction}, increasing model capacity generally improves performance, with the 44.97M parameter \modelname~ variant often achieving the best results over the MolGPS baseline. The impact of incorporating MoA information varies significantly across datasets: while it enhances performance on MoAT-DB validation and OOD test sets (improving ACC@1\% from 0.560 to 0.626 on $\mathcal{D}_\text{MoA}$), it shows mixed results on external benchmarks. In particular, performance degrades on KIBA and DAVIS datasets, where inhibitor annotations were broadly assigned, often based on binding affinity or computational heuristics rather than experimental confirmation. These results demonstrate that while model capacity consistently benefits prediction, the utility of additional biological context depends heavily on the quality of the prior.

\clearpage

\end{document}